\pdfoutput=1
\documentclass[letterpaper, 10 pt, journal, twoside]{IEEEtran}

\usepackage{times}
\usepackage[pdftex]{graphicx}
\usepackage{subfigure}
\usepackage{amsmath,amssymb,amsopn,amstext,amsfonts}
\usepackage{cancel}
\usepackage[space]{cite}
\usepackage{soul}
\usepackage{cuted}
\usepackage{lipsum,multicol}
\usepackage{stfloats} 
\everymath{\displaystyle}
\usepackage{booktabs}
\usepackage{threeparttable}
\usepackage{amsthm}
\usepackage{xfrac}
\usepackage{balance}
\usepackage{color}
\usepackage{mathtools}

\usepackage{tabularx}
\usepackage{bm}
\usepackage{diagbox}
\usepackage{float}
\usepackage{epstopdf}
\usepackage{url}
\usepackage{multirow}
\usepackage{multicol}
\usepackage{tikz}
\usepackage[linkcolor=black,citecolor=black,urlcolor=black,colorlinks=true]{hyperref}
\usepackage{enumitem}
\usepackage[ruled,vlined,linesnumbered]{algorithm2e}
\usepackage{pifont}
\usepackage{siunitx}
\usepackage{hyperref}
\usepackage{colortbl, array}
\usepackage{booktabs}

\DeclareMathAlphabet\mathbfcal{OMS}{cmsy}{b}{n}

\let\oldFootnote\footnote
\newcommand\nextToken\relax

\renewcommand\footnote[1]{%
    \oldFootnote{#1}\futurelet\nextToken\isFootnote}

\newcommand\isFootnote{%
    \ifx\footnote\nextToken\textsuperscript{,}\fi}

\definecolor{celadon}{rgb}{0.78, 0.93, 0.80}
\pagecolor{celadon}
\nopagecolor
\soulregister\cite7
\soulregister\citep7
\soulregister\citet7
\soulregister\ref7
\soulregister\it7
\soulregister\pageref7

\usepackage{titlesec}

\renewcommand\thesubsubsection{\arabic{subsubsection})}
\renewcommand\theparagraph{\alph{paragraph})}

\makeatletter
\@addtoreset{paragraph}{subsubsection}
\makeatother

\titleformat{\paragraph}[runin]
  {\normalfont\normalsize}
  {\theparagraph}{0.5em}{}[]

\usepackage{arydshln}
\graphicspath{{figures/}}
\DeclareGraphicsExtensions{.pdf,.png,.jpg,.eps}

\IEEEoverridecommandlockouts

\title{
A Survey on LiDAR-based Autonomous \\ Aerial Vehicles
}

\author{Yunfan Ren, Yixi Cai, Haotian Li, Nan Chen, Fangcheng Zhu, \\Longji Yin, 
Fanze Kong, Rundong Li and Fu Zhang
    \thanks{Corresponding author: Fu Zhang.}    
    \thanks{All authors are with Mechatronics and Robotic Systems (MaRS) Laboratory, Department of Mechanical Engineering, University of Hong Kong, Hong Kong SAR, China. (email: \{renyf, yixicai\}@connect.hku.hk; fuzhang@hku.hk)}
}

\begin{document}
\maketitle

\begin{tikzpicture}[overlay, remember picture]
    \path (current page.north) ++(0.0,-1.0) node[draw = black] {This article has been accepted by IEEE/ASME Transactions on Mechatronics, 2025};
\end{tikzpicture}
\vspace{-0.3cm}

\begin{abstract}

    This survey offers a comprehensive overview of recent advancements in LiDAR-based autonomous Unmanned Aerial Vehicles (UAVs), covering their design, perception, planning, and control strategies. Over the past decade, LiDAR technology has become a crucial enabler for high-speed, agile, and reliable UAV navigation, especially in GPS-denied environments.
    The paper begins by examining the evolution of LiDAR sensors, emphasizing their unique advantages such as high accuracy, long-range depth measurements, and robust performance under various lighting conditions, making them particularly well-suited for UAV applications.
    The integration of LiDAR with UAVs has significantly enhanced their autonomy, enabling complex missions in diverse and challenging environments.
    Subsequently, we explore essential software components, including perception technologies for state estimation and mapping, as well as trajectory planning and control methodologies,
    {and discuss their adoption in LiDAR-based UAVs.
            Additionally, we analyze various practical applications {of the LiDAR-based UAVs}, ranging from industrial operations to supporting different aerial platforms and UAV swarm deployments.}
    The survey concludes by {discussing} existing challenges and proposing future research directions to advance {LiDAR-based} UAVs and enhance multi-UAV collaboration. By synthesizing recent developments, this paper aims to provide a valuable resource for researchers and practitioners working to push the boundaries of LiDAR-based UAV systems.
\end{abstract}

\begin{IEEEkeywords}
    LiDAR-based UAV, Applications on UAV, UAV Autonomy
\end{IEEEkeywords}

\section{Introduction}
\label{sec:introduction}

The past decade has witnessed a remarkable surge in the development and deployment of Unmanned Aerial Vehicles (UAVs), driven by their versatility, cost-effectiveness, and ability to access hazardous or hard-to-reach areas.
These aerial systems have found widespread applications in fields ranging from aerial photography~\cite{kovanivc2023review}, precision agriculture~\cite{tsouros2019review} to infrastructure inspection~\cite{jordan2018state}, search and rescue operations~\cite{daud2022applications,rabta2018drone}.
By endowing these aerial platforms with the ability to sense, make decisions and navigate without direct human control, the scope of their applications has expanded significantly. Autonomous UAVs can execute intricate missions, such as surveying remote or hazardous areas, without exposing human operators to unnecessary risks.

\begin{figure}[t]
    \centering
    \includegraphics[width=0.48\textwidth]{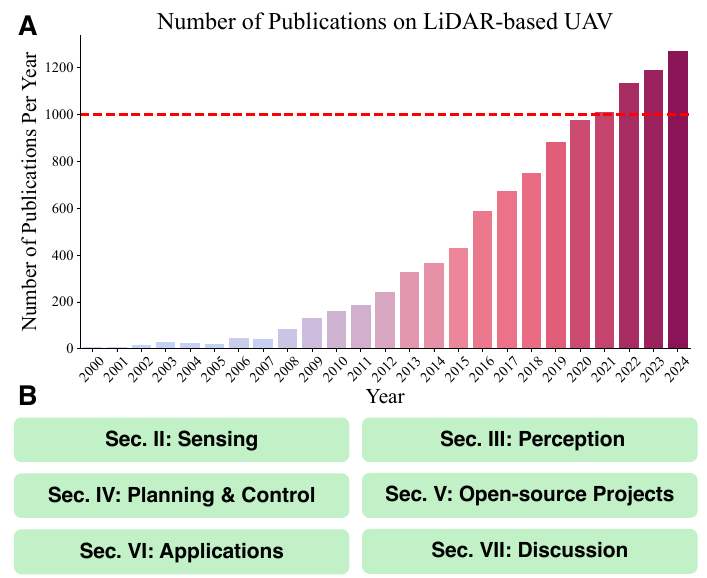}
    \caption{(\textbf{A}) Number of publications on LiDAR-based UAV since 2000. In {2021}, the number of publications related to LiDAR-based UAV surpassed one thousand.
        {These data were retrieved from the Web of Science,
            using the search terms: (``LiDAR'' OR ``Laser'') AND (``UAV'' OR ``MAV'' OR ``Drone'' OR ``Aerial'')}.
    {(\textbf{B}) The architecture of this paper.}
    }
    \label{fig:pub_num}
    \vspace{-0.6cm}
\end{figure}

For autonomous navigation, the sensing module is essential for UAVs to achieve reliable state estimation and obstacle detection.
    {Radar, popular in autonomous driving~\cite{abu2023radar}, excels in adverse weather and offers long-range detection with low weight and power needs. Vision-based cameras, widely used in UAVs~\cite{shen2013vision,oleynikova2018safe,zhou2019robust}, leverage size, weight, and power (SWaP) advantages but struggle with depth accuracy, motion blur, and poor lighting. Despite being larger and heavier, LiDAR remains highly appealing for UAVs due to its unique strengths. It provides active, centimeter-level depth measurements over hundreds of meters, enabling high-speed autonomous flight (e.g., 20 m/s~\cite{ren2025safety}) and navigation in cluttered environments~\cite{ren2023online}. Operating at high frequencies (hundreds of thousands to millions of Hertz), it generates dense 3D point clouds for fast, accurate motion estimation~\cite{xu2022fast,he2023point}. Its time-of-flight (ToF) mechanism also excels at detecting thin objects, such as power lines, which cameras often miss due to limited resolution or contrast, and radar may overlook due to lower spatial resolution and material-dependent reflectivity. Compared to radar, LiDAR offers exceptional resolution and reliability in complex, obstacle-laden environments, making it well-suited for the dynamic, three-dimensional maneuvers demanded by UAVs. These strengths overshadow LiDAR’s size and weight disadvantages, particularly as ongoing progress in miniaturization and lightweight design steadily mitigates these constraints.}

    {LiDAR technology is widely utilized across various robotic platforms, including wheeled robots~\cite{jian2023dynamic} and quadruped robots~\cite{lee2020learning}. However, this paper focuses specifically on its application in unmanned aerial vehicles (UAVs) due to the unique challenges they encounter. Unlike ground-based platforms, UAVs operate in dynamic, three-dimensional environments while contending with constraints such as limited payload capacity, strict power efficiency requirements, and the need for real-time navigation in complex, obstacle-dense settings (e.g., urban landscapes or forests). These conditions impose distinct demands on LiDAR systems, encompassing both hardware and software considerations. For instance, UAV-mounted LiDAR must be lightweight, capable of high-resolution 3D mapping, and supported by perception algorithms that are robust against vibrations (e.g., those caused by rotors and propellers) as well as altitude fluctuations. In contrast, wheeled and quadruped robots typically operate in more stable, two- or 2.5-dimensional environments, where weight and power constraints are less restrictive, and the impact of vibrations is significantly reduced.}


Since the 1990s, research on autonomous UAVs has gained increasing popularity. {Concurrently,} the development of LiDAR sensors significantly enhancing UAV autonomy. This advancement has led to the emergence of various techniques and the publication of over a thousand papers annually related to LiDAR-based UAV systems, as shown in Fig.~\ref{fig:pub_num}. Several {existing} survey papers have reviewed autonomous UAVs, such as Kendoul's comprehensive survey on UAV guidance, navigation, and control in 2012\cite{kendoul2012survey}, including both vision-based and LiDAR-based systems; Kanellakis \textit{et al.}'s review of vision-based works~\cite{kanellakis2017survey}; and reviews by Gyagenda \textit{et al.}\cite{gyagenda2022review} and Chang \textit{et al.}\cite{chang2023review} on UAV navigation techniques in GNSS-denied environments. { Additionally, surveys have specifically focused on LiDAR-based perception technologies, such as Lee \textit{et al.}'s extensive review of LiDAR odometry across robotics~\cite{lee2024lidar} and Zhang \textit{et al.}'s focus on LiDAR-based place recognition for autonomous driving~\cite{zhang2024lidar}. However, despite these efforts, a dedicated and comprehensive review of LiDAR-based autonomous UAV systems remains absent.}

    {This survey aims to fill that gap by addressing the lack of a specialized review on LiDAR-based autonomous UAV systems, particularly given the suitability of LiDAR sensors for enabling aggressive and high-speed UAV maneuvers. }
The primary objective of this paper is to provide an overview of the history, state-of-the-art advancements, milestones, and unresolved challenges in the field of LiDAR-based autonomous UAVs. The rest of the paper is organized as follows: Sec.~\ref{sec:dev} introduces the advantages and development of LiDAR sensors and LiDAR-enabled autonomous UAVs. Sec.~\ref{sec:perception} presents LiDAR-based perception technologies suitable for UAV applications. In Sec.~\ref{sec:pac}, we discuss planning and control technologies that leverage LiDAR to enable higher levels of UAV autonomy. Sec.~\ref{sec:open} reviews existing open-source modules useful for building LiDAR-based autonomous UAVs from the ground up. Sec.~\ref{sec:app} explores various applications of LiDAR-based UAVs. Finally, Sec.~\ref{sec:disc} provides discussions and possible future directions for LiDAR-based autonomous UAVs.

\begin{figure*}[t]
    \centering
    \includegraphics[width=0.95\textwidth]{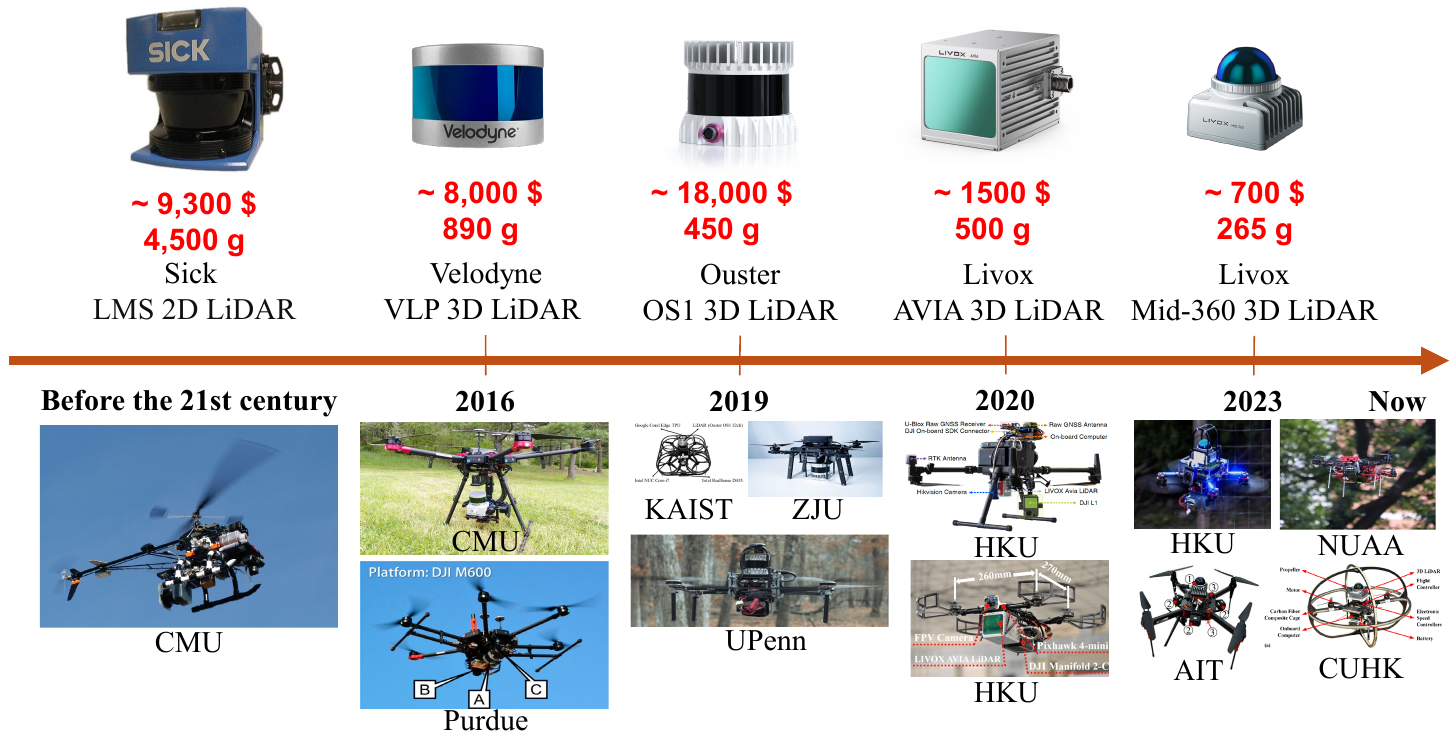}
    \caption{
        {Overview of LiDAR developments and corresponding UAV demonstrations in recent years, including Sick LMS 2D LiDAR\cite{thrun2006scan}, Velodyne VLP 3D LiDAR\cite{zhang2018p,lin2019evaluation}, Ouster OS1 3D LiDAR\cite{wang2022geometry,cheng2024treescope,kim2024autonomous}, Livox AVIA solid-state 3D LiDAR\cite{kong2021avoiding,li2024mars}, and Livox MID-360 solid-state 3D LiDAR\cite{ren2022bubble,liu2025slope,ma2024research,bolz2024robust,wang2024air,zhang2024developing}}
    }
    \label{fig:li_dev}
    \vspace{-0.5cm}
\end{figure*}

\section{Development of LiDAR Sensors and LiDAR-based UAVs}
\label{sec:dev}

{LiDAR, first deployed in the 1960s for satellite tracking~\cite{stitch1961optical}, operates on the time-of-flight (ToF) principle: laser pulses are emitted, and their return times are measured to calculate distances with centimeter-level precision. Unlike radar, which uses radio waves for long-range detection (kilometers) and performs well in adverse weather due to its longer wavelengths, LiDAR relies on shorter-wavelength light (typically 905 nm or 1550 nm). This allows for superior spatial resolution but makes it susceptible to scattering in heavy rain or fog. Compared to cameras, which produce 2D pixel arrays rich in color and texture but lack direct depth perception, LiDAR generates 3D point clouds—sets of (x, y, z) coordinates—that provide precise geometry ideal for mapping complex environments.}

{The development of LiDAR-based UAVs has closely mirrored advances in LiDAR technology, progressing from bulky, high-cost systems to compact, affordable sensors that support real-time autonomous flight. Initial UAV integrations employed 2D LiDARs mounted on large unmanned helicopters—typically 3 meters long and weighing 60–90 kg—for tasks such as mapping~\cite{miller19983}, obstacle avoidance~\cite{shim2006conflict}, and autonomous landing~\cite{theodore2006flight}. However, the size and cost of early LiDAR systems restricted broader adoption.}

{A significant breakthrough occurred with the introduction of the Velodyne HDL-64E, which was used in the 2007 DARPA Urban Challenge~\cite{bohren2008little}. This sensor offered 64 beams, 360° coverage, and centimeter-level accuracy, setting a new performance benchmark. Nevertheless, its substantial weight (8 kg) and high cost (USD 8,000–20,000) confined its use to larger platforms. The 2015 release of the Velodyne VLP-16 marked a turning point, as it delivered 16 beams and a 100–200 m range in a much smaller (900 g) and more affordable (USD 4,000) package. This enabled integration into smaller UAVs, with Gao et al.\cite{gao2016online} demonstrating its use in onboard state estimation and obstacle avoidance, and Zhang et al.\cite{zhang2020falco} achieving autonomous flight at speeds of up to 10 m/s in unknown environments.
    Subsequent advances brought high-resolution mechanical LiDARs such as the Ouster OS1-128~\cite{ouster128}, which features 128 beams and dense point clouds at a reduced cost (USD 5,000–12,000)\cite{wang2022geometry}. These improvements were largely driven by component standardization, enhanced semiconductor manufacturing, and economies of scale.}

{More recently, compact solid-state LiDARs have further expanded UAV capabilities. Devices like the Livox Avia (500 g, USD 1,500\cite{avia}) and the MID360 (265 g, USD 700~\cite{manual2023mid360}) employ non-mechanical scanning mechanisms, such as Risley prisms, to achieve 70–200 m range with centimeter-level accuracy. These sensors are particularly well suited for small multirotor UAVs operating in cluttered or GNSS-denied environments~\cite{ren2023online}, and have been shown to support high-speed flight exceeding 20 m/s~\cite{ren2025safety}. Emerging technologies such as frequency-modulated continuous wave (FMCW) LiDAR offer 100–300 m range, direct velocity measurement, and resilience to fog and interference~\cite{kim2020fmcw, sayyah2022fully}. However, current pricing (USD 1,500–10,000) remains a barrier to widespread UAV adoption.}

{The evolution of LiDAR technology for UAV applications—alongside its cost reduction—has been driven by three key trends: the transition from mechanical to solid-state architectures, the integration of on-chip processing, and the adoption of scalable manufacturing through component standardization. As a result, different LiDAR types now support a broad spectrum of UAV missions. Mechanical LiDARs, such as those from Velodyne and Ouster, remain well-suited for high-resolution mapping on larger platforms. Solid-state LiDARs, including Livox and MID360, provide lightweight, low-SWaP solutions ideal for agile navigation in complex environments. Meanwhile, FMCW LiDARs offer strong potential for long-range, high-speed, and weather-resilient operations, though current costs still limit their deployment.}

{A typical autonomous system includes hardware for sensing, computation, and actuation, as well as software for perception, planning, and control. An ideal sensor suite should offer comprehensive environmental information under various conditions (e.g., different lighting) and enable robust, real-time ego-state estimation. LiDAR plays a critical role in UAV systems, particularly in perception, where its high-accuracy geometry, high point rate, and reliable performance in low-light or featureless environments support robust state estimation~\cite{xu2022fast, he2023point} and precise mapping~\cite{ren2024rog, cai2023occupancy}. In planning, LiDAR’s long-range obstacle detection aids high-speed navigation (e.g., 20 m/s)\cite{ren2025safety}, while its precision enables accurate navigation in confined spaces\cite{ren2023online}. The accurate state estimation based on LiDAR also enhances control systems, allowing for precise tracking even during aggressive maneuvers~\cite{mpc_luguozheng}. Despite its limitations—such as lack of color information and susceptibility to weather-related signal scattering—ongoing advancements in miniaturization and cost reduction continue to improve its suitability for UAVs, as shown in Fig.~\ref{fig:li_dev}.}

\section{LiDAR-based Perception}
\label{sec:perception}

\subsection{State Estimation for UAVs}

\begin{figure}[htbp]
    \vspace{-0.2cm}
    \centering
    \includegraphics[width=0.48\textwidth]{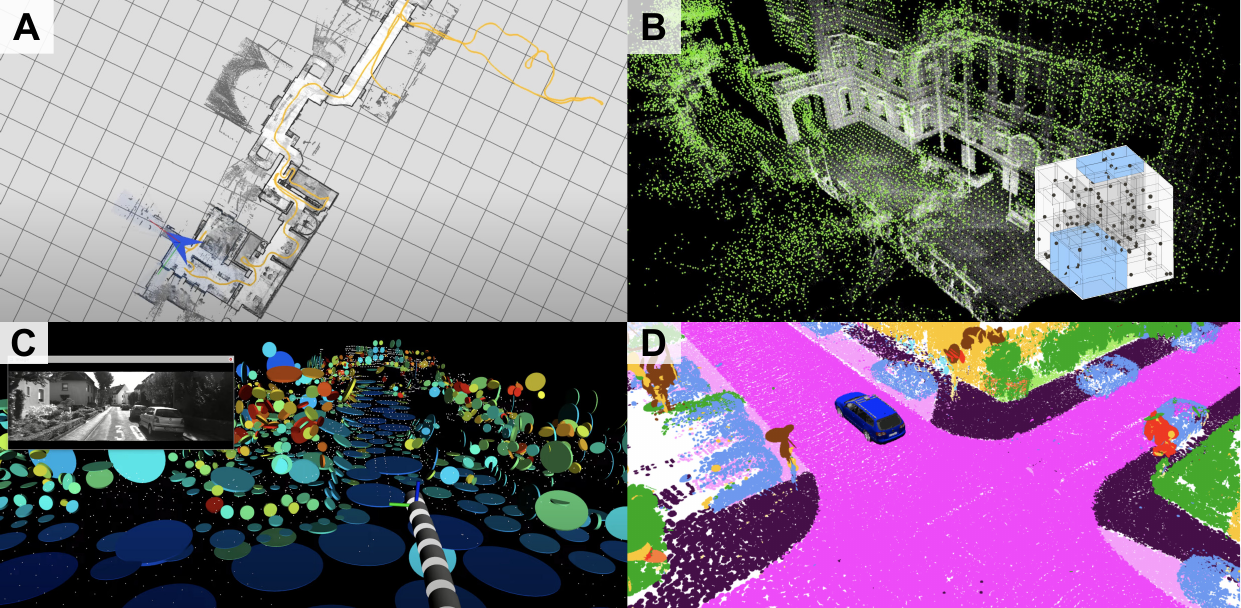}
    \caption{{Illustration of four notable state estimation methods: (\textbf{A}) Example of 2D LiDAR SLAM~\cite{hess2016real}; (\textbf{B}) Nearest neighbor search-based (incremental) KD-tree approach, as seen in~\cite{cai2021ikd, xu2022fast, zhu2022robust, zhu2023swarm, zhu2024swarm, he2023point}; (\textbf{C}) Adaptive probabilistic voxel map-based state estimation~\cite{yuan2022efficient,zheng2022fast,zheng2024fast}; (\textbf{D}) State estimation using a semantic map~\cite{chen2019suma++,chen2020sloam}.}
    }
    \vspace{-0.3cm}
    \label{fig:state_est}
\end{figure}

{For an autonomous robot, the first question it must address is its own location. This is the primary role of the state estimation module.
    State estimation aims to accurately determine the UAV’s state, including attitude, position, and velocity, by employing algorithms that estimate the state from LiDAR point measurements (e.g., \cite{zhang2014loam}) or integrate LiDAR data with other sensors, such as an IMU~\cite{xu2022fast} or a camera~\cite{zheng2024fast}.} It serves as the foundation for autonomous UAV systems, supporting upper-layer modules such as planning and control. LiDARs are widely used for high-quality state estimation due to their unique advantages, including long-range detection, precise distance measurements, and robustness to varying lighting conditions. As LiDAR-based state estimation is closely tied to advancements in LiDAR technology, we separately discuss algorithms using 2D and 3D LiDARs {and offer a concise summary of key technologies in Fig.~\ref{fig:state_est}}.

\subsubsection{2D LiDAR-based State Estimation}
While UAVs equipped with LiDAR (including 2D laser range finders) emerged in the late 20th century~\cite{miller19983}, LiDAR-based state estimation for UAVs gained momentum in the early 21st century~\cite{thrun2006scan, theodore2006flight, angeletti2008autonomous,he2008planning,he2010efficient, bry2012state}. In 1998, Miller \textit{et al.} used a 2D laser range finder for environmental mapping, but due to the lack of reliable localization algorithms, the state estimation relied on GPS, IMU, and compass sensors.
By 2003, Thrun \textit{et al.}\cite{thrun2006scan} implemented a probabilistic SLAM framework using 2D LiDAR, GPS, and compass data on a helicopter, employing the Iterative Closest Point (ICP) algorithm\cite{besl1992method} for reliable pose estimation. In 2006, Theodore \textit{et al.}~\cite{theodore2006flight} combined 2D LiDAR with monocular camera images and IMU data for precise landing, using LiDAR to ensure accurate positioning near the ground.
Initially, 2D LiDARs were auxiliary to GPS, suitable only for outdoor environments with strong GPS signals. However, GPS-denied scenarios, such as indoor environments, demanded alternative solutions. In 2008, Angeletti \textit{et al.}\cite{angeletti2008autonomous} achieved indoor autonomous hovering by matching 2D LiDAR scans to estimate horizontal position and yaw angle, while ultrasonic sensors and IMU provided vertical position and attitude data for 6-DOF pose estimation. That same year, He \textit{et al.}\cite{he2008planning} used a redirected 2D LiDAR scan for downward range measurements, enabling geometric beacon tracking to achieve GPS-free autonomous flight.
In 2012, Bry~\cite{bry2012state} introduced a hybrid filter that tightly coupled IMU and 2D LiDAR data for localization on a prior map. The approach used an IMU-driven Extended Kalman Filter (EKF) for prediction, with Gaussian Particle Filter (GPF) updates, resulting in robust and efficient state estimation. {Later in 2016, Hess \textit{et al.} introduced the loop closure technique into a real-time 2D LiDAR SLAM system on a portable platform~\cite{hess2016real}.}

\subsubsection{3D LiDAR-based State Estimation}
While 2D LiDARs offer precise distance measurements, their limited vertical FoV restricts their effectiveness for UAVs with 3D motion. The introduction of the VLP-16 in 2015~\cite{vlp16} marked a major milestone in LiDAR development, quickly gaining popularity in fields such as autonomous driving, robotics, and UAV technology.

One of the earliest and most influential 3D LiDAR SLAM algorithms, LOAM, was introduced by Zhang \textit{et al.} in 2014~\cite{zhang2014loam}. Unlike traditional point-to-point ICP methods, LOAM extracts plane and edge features from point clouds and optimizes state estimation through point-to-plane and point-to-edge ICP {method}.
This framework has been widely adopted in UAV applications~\cite{gao2016online, liu2017planning, gao2019flying}, with later enhancements integrating IMU and cameras for robust performance during aggressive UAV maneuvers~\cite{zhang2018pcap, zhang2019pcal, zhang2020falco}. {In addition to the point cloud, semantic information is leveraged to enhance performance~\cite{chen2020sloam,chen2019suma++}. Furthermore, instead of directly utilizing LiDAR points in the map, Behley \textit{et al.}~\cite{behley2018efficient} employed a surfel representation for plane features in LiDAR SLAM. In 2019, Ye \textit{et al.} introduced IMU into the LiDAR SLAM system to reduce the drift in long-term mapping\cite{ye2019tightly}. To achieve higher robustness, Jiao \textit{et al.} proposed M-LOAM~\cite{jiao2021robust}, integrating measurements from multiple LiDARs to achieve accuracy and consistent odometry and mapping. }

The emergence of solid-state LiDARs {\cite{liu2021low}} like Livox Avia and Mid360~\cite{avia,manual2023mid360} in the 2020s further accelerated the development of LiDAR-based UAV state estimation. Many innovative algorithms have since emerged~\cite{lin2020loam,xu2021fast,xu2022fast,he2023point,zhu2023swarm,zhu2024swarm}.
In 2020, Lin \textit{et al.}\cite{lin2020loam} proposed LOAM-Livox, which pioneered a LiDAR odometry system for small field-of-view solid-state LiDARs by introducing a scan-to-map registration approach. In 2021, Xu \textit{et al.}\cite{xu2021fast} introduced FAST-LIO, an efficient LiDAR-inertial odometry framework based on ESIKF, featuring a new formula for Kalman gain calculation to manage large-scale point clouds and enhance computational efficiency. FAST-LIO2\cite{xu2022fast} extended this framework with the ikd-tree~\cite{cai2021ikd}, optimizing nearest-neighbor search and enabling direct point registration, improving both robustness and efficiency. These advancements have been widely adopted in UAV research~\cite{ren2022bubble,kong2021avoiding,ren2023online,tang2023bubble,ren2024rog,cai2023occupancy}.
He \textit{et al.}~\cite{he2023point} further enhanced the FAST-LIO framework by implementing point-wise state estimation, dramatically increasing output frequency and enabling robust performance during extreme maneuvers, even with saturated IMU measurements.

When utilizing these LIO methods as the state estimation module of UAV systems, accurate temporal-spatial calibration between LiDAR and IMU sensors is always required. To solve this problem, Zhu \textit{et al.}~\cite{zhu2022robust} proposed a real-time initialization framework {that precisely estimates the temporal-spatial extrinsics between LiDAR and IMU sensors.} {Liu \textit{et al.}~\cite{liu2023time} proposed a time delay error correction method to online estimate the time offset between Lidar and IMU sensors.}

The camera is an excellent complementary sensor to 3D LiDAR due to its low cost and rich color information. In recent years, many state estimation methods that integrate LiDAR, camera, and IMU sensors have emerged~\cite{zheng2022fast, zheng2024fast, lin2024r, lv2023continuous}. {In 2022, Zheng \textit{et al.} proposed FAST-LIVO\cite{zheng2022fast}, which integrates the strengths of sparse direct image alignment and direct raw LiDAR point registration to deliver accurate and robust pose estimation. Building on this foundation, FAST-LIVO2 \cite{zheng2024fast} was presented in 2024, offering improvements in computational efficiency, memory management, localization accuracy, and overall robustness.} This method was successfully applied to UAV systems, enabling autonomous navigation.

    {Expanding on these advancements, recent research has also explored applications in aerial swarm systems, such as} Zhu \textit{et al.}\cite{zhu2023swarm,zhu2024swarm} introduced a fast LiDAR-inertial odometry framework optimized for swarm systems using Livox LiDARs, enabling real-time ego and mutual state estimation for collaborative missions\cite{yin2023decentralized}. In 2023, Pritzl \textit{et al.}~\cite{pritzl2023fusion} demonstrated the use of 3D LiDAR for relative localization, effectively mitigating visual-inertial odometry drift through LiDAR measurements.

\subsubsection{Discussion}
The advancement of LiDAR-based state estimation methods is closely tied to the development of LiDAR sensor technology. Compared to 2D LiDARs, 3D spinning LiDARs offer a wider FoV, while the advent of solid-state LiDARs introduces benefits like reduced cost and lighter weight. Each sensor innovation has spurred new state estimation algorithms, expanding LiDAR applications from ground robots to UAVs and even aerial swarms. As LiDAR technology evolves, emerging sensors like Frequency Modulated Continuous Wave (FMCW) LiDAR, which provides velocity measurements for each point, are expected to inspire more advanced algorithms, further advancing autonomous drone capabilities.

Due to motion distortion and performance degradation in low-structured environments, most LiDAR-based state estimation methods rely on multi-sensor fusion, integrating data from IMU, GNSS, and other sensors. This trend is expected to continue, with future systems incorporating additional sensors, such as cameras and radar, to enhance state estimation accuracy and robustness.

\subsection{Occupancy Mapping}
{To enable a UAV to navigate effectively in an unknown environment, it must possess an understanding of the space’s traversability. Occupancy mapping, a form of volumetric mapping, addresses this challenge by constructing consistent maps that categorize regions as occupied, free, or unknown, thereby facilitating safe and efficient navigation for UAVs in unexplored settings~\cite{thrun2002probabilistic}}. This section discusses discrete and continuous occupancy representations, followed by an overview of popular map structures used to manage occupancy information.
\subsubsection{Occupancy Rerepsentations}
Discrete representations are popular in occupancy mapping for their simplicity and efficiency. A common approach divides the space into evenly distributed 2D grids or 3D voxels, each represented by a Bernoulli random variable indicating occupancy probability. By assuming spatial independence between neighboring voxels, occupancy probabilities can be efficiently updated. This technique, known as occupancy grid mapping, was introduced in early research~\cite{elfes1987sonar, moravec1988sensor} and formalized in~\cite{thrun2002probabilistic}. Octomap~\cite{hornung2013octomap} extended this framework to 3D using octrees, {which uniformly partition the environment into octants}, and a log-odd function to reduce computational complexity.

Another discrete method is the Euclidean Signed Distance Field (ESDF), which computes the Euclidean distance from each voxel to the nearest surface. While most ESDF representations are derived from occupancy grids, Voxblox~\cite{oleynikova2017voxblox} constructs ESDF maps directly from Truncated Signed Distance Field (TSDF) data. However, Voxblox suffers from distance estimation inaccuracies due to the approximated TSDF-to-ESDF conversion~\cite{han2019fiesta}.

Continuous representations assume a spatial correlation between neighboring spaces, reflecting real-world physical structures. Early attempts~\cite{veeck2004learning, paskin2012robotic} modeled boundaries and occupancy probabilities using curves and polygons. Later, O'Callaghan and Ramos~\cite{o2012gaussian} proposed a non-parametric approach using Gaussian processes (GPs) to estimate continuous occupancy in 2D space. However, GPs require storing all $N$ measurements and performing $\boldsymbol{\mathcal{O}}(N^3)$ operations for each query, making them computationally expensive. Although clustering local measurements can reduce the dataset size, each query still requires constructing a new covariance matrix. Extensions like GPmap~\cite{kim2015gpmap}, GPOctomap~\cite{wang2016fast}, and BGKOctomap-L~\cite{doherty2019BGK} applied GPs to 3D spaces using octrees~\cite{jackins1980octree}, but they continued to rely on raw measurements for predictions.

Semi-parametric methods mitigate the storage burden by clustering data into parameterized functions, such as Gaussians or kernels. NDT-OM\cite{saarinen20133d} divided space into voxels, with each voxel storing a Gaussian, though it struggled with discretization errors in unstructured environments. Confidence-rich mapping\cite{agha2019confidence} accounted for measurement uncertainty within sensor cones to improve planning and exploration. The Hilbert map\cite{hilbertmap} and its variants\cite{incrementalhilbertmap, guizilini2016large, zhi2019continuous} employed kernel approximations to reduce inference complexity to $\boldsymbol{\mathcal{O}}(N)$. Additionally, occupancy maps based on Gaussian mixture models have been explored, with hierarchical structures~\cite{srivastava2018efficient} and variable resolutions~\cite{o2018variable} providing further flexibility.

\subsubsection{Occupancy Map Structures}
As discrete representations divide the space into evenly distributed 2D cells or 3D voxels, the early research utilized arrays to store grids, typically known as uniform grid maps~\cite{fankhauser2016universal}. However, a critical drawback of the uniform grid maps is their tremendous memory consumption, which prevents their application in high-resolution and large-scale mapping. Nonetheless, uniform grid maps offer the strongest efficiency in updating and querying compared to other map structures to be discussed in this section, due to their continuous memory allocation. Thus, uniform grid maps are well-suited for occupancy mapping in a local space, {such as the ROG-Map~\cite{ren2024rog}}.

To reduce the memory consumption in uniform grid maps, hashing techniques were introduced to organize the cells (or voxels)~\cite{niessner2013hashmap}. Rather than pre-allocating memory for each cell (or voxel), the hashing-based grid map allocates a smaller array.
The hashing grid map allows dynamic map resizing through rehashing and reallocation, thus not requiring knowledge of the mapping area beforehand~\cite{oleynikova2017voxblox}. Compared to uniform grid maps, hashing grid maps possess higher memory efficiency and better dynamic ability. However, the computational efficiency of hashing grid maps is lower than that of uniform grid maps due to hash conflicts and a lower cache hit rate~\cite{ericson2004real}.

Quadtrees and octrees are useful data structures for organizing voxels at various resolutions and have become popularly used in occupancy grid mapping~\cite{kraetzschmar2004probabilistic, hornung2013octomap}. These tree-based map structures exhibit superior memory efficiency compared to uniform grid maps and hashing grid maps, making them favorable for high-resolution maps and large-scale environments. However, the time complexity for updating occupancy probabilities in a tree structure is logarithmic, while that in grid maps is constant. Consequently, subsequent research has focused on enhancing update efficiency without compromising original accuracy and memory efficiency~\cite{wurm2011hierarchies,vespa2018efficient,duberg2020ufomap,funk2021multi}. One of the notable research studies for LiDAR sensors is D-Map\cite{cai2023occupancy}. Instead of relying on traditional ray casting to determine occupancy states, D-Map leverages depth image projection and an on-tree update strategy to achieve high-efficiency performance, while also capitalizes on the high accuracy of LiDAR sensors to further reduce the computational burden.

The occupancy map structures for continuous representation are similar to those for discrete representations. Grid maps and tree-based maps are used to store raw measurements~\cite{kim2015gpmap, wang2016fast, doherty2019BGK} or parameterized clusters of raw measurements~\cite{saarinen20133d,agha2019confidence,srivastava2018efficient,o2018variable}. Another data structure worth mentioning is the R-tree~\cite{guttman1984rtree} which is used to organize GMMs without evenly partitioning the space, thus avoiding discretization errors~\cite{li2024gmmap}.

\subsubsection{Discussion}
When comparing continuous representations with discrete ones, continuous occupancy mapping approaches effectively consider spatial correlation and provide mapping with infinite resolution as well as uncertainty estimation. However, a notable limitation lies in their computational efficiency, which currently hinders their practical application in real robotic systems, especially in aerial systems with limited computational resources. Consequently, discrete representations are currently preferred, but the full potential of continuous mapping could be unlocked with advanced onboard computational resources and the exploration of new designs in map structures utilizing parallel computation architectures.

\subsection{Dynamic Object Detection and Tracking}

{
    In the final part of this section, we briefly overview key technologies in LiDAR-based dynamic object detection and tracking. These approaches enable UAVs to avoid dynamic obstacles and navigate crowded environments effectively.
}

{Model-based methods harness the geometric properties of point clouds. For example, Dewan \textit{et al.} introduced a model-free approach in 2016 that uses motion cues and Bayesian segmentation to predict dynamic objects as bounding boxes, addressing partial observation and occlusion challenges \cite{dewan2016motion}. Yoon \textit{et al.} developed a mapless online detection method that corrects motion distortion and classifies dynamic points through freespace querying \cite{yoon2019mapless}. Duberg \textit{et al.} presented DUFOMap, an efficient mapping framework that identifies dynamic regions by detecting void spaces, delivering high accuracy and computational efficiency~\cite{duberg2024dufomap}. Wu \textit{et al.} proposed M-detector, a microsecond-latency method that detects moving events in LiDAR streams based on occlusion principles, achieving excellent accuracy and efficiency \cite{wu2024moving}. Lu \textit{et al.} introduced a geometric segmentation method with an adaptive-covariance Kalman filter, enabling real-time tracking and collision avoidance on a quadrotor UAV \cite{lu2024fapp}.}

{Learning-based methods also shine. Mersch \textit{et al.} introduced a sparse 4D convolution-based approach for online segmentation of moving objects in 3D LiDAR data, delivering SOTA results on SemanticKITTI and generalizing to unseen environments like Apollo \cite{mersch2022receding}. Zhang \textit{et al.} developed DeFlow, a real-time scene flow estimation network using GRUs to extract point-specific features from voxelized data, tackling data imbalance with a novel loss function \cite{zhang2024deflow}. Together, these advancements address challenges like motion distortion, real-time processing, and adaptability, advancing LiDAR-based dynamic perception for autonomous systems. }

\section{Planning and Control}
\label{sec:pac}

{LiDAR-based perception provides UAVs with high-resolution 3D maps, offering centimeter-level accuracy and extended sensing ranges (tens to hundreds of meters), far surpassing cameras like the Intel D435i~\cite{d435i} (3–5 m). While planning and control modules are generally sensor-agnostic post-mapping, LiDAR’s superior range and accuracy introduce unique challenges and opportunities. For planning, existing vision-based methods can leverage LiDAR’s detailed maps directly, but to fully exploit UAV maneuverability, algorithms must efficiently handle longer planning horizons. Additionally, LiDAR’s robust, accurate state estimation enables agile, high-speed maneuvers, requiring planning modules to generate aggressive trajectories and control modules to utilize low-latency, precise data for enhanced performance. This distinguishes LiDAR-based systems from camera-based approaches, justifying advanced design considerations. In this section, we review existing planning and control methods, then discuss their adaptation for LiDAR-based UAVs to enable agile, high-speed autonomous flight in complex environments.}

\subsection{Trajectory Planning}

\begin{figure}[t]
    \centering
    \includegraphics[width=0.48\textwidth]{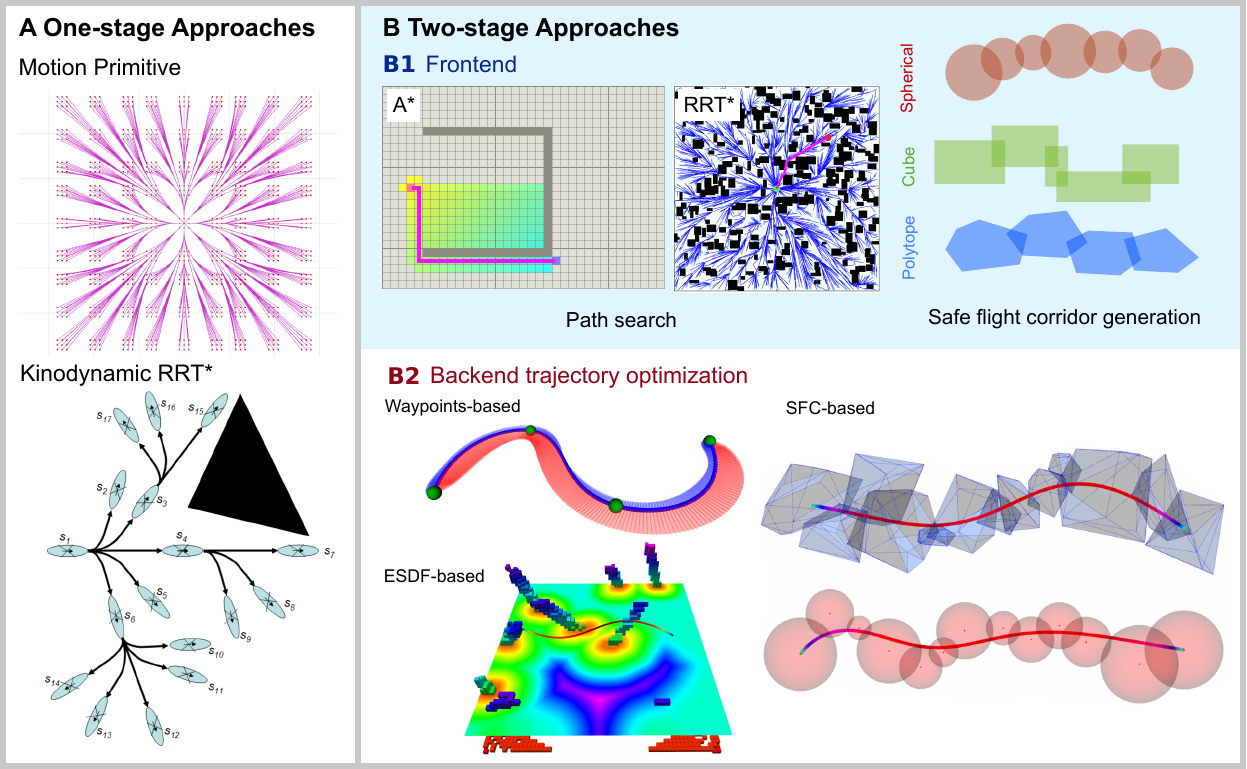}
    \caption{Existing planning approaches can be categorized into two types: (\textbf{A}) One-stage approaches directly generate a collision-free, dynamically feasible trajectory from the current state to the goal. (\textbf{B}) Two-stage approaches involve a front-end path planning module and a back-end trajectory optimization module.
    }
    \label{fig:planning}
    \vspace{-0.45cm}
\end{figure}

\label{sec:tp}
The trajectory planning module generates safe, collision-free, and dynamically feasible paths toward mission goals based on the mapping results and the UAV’s current state, while minimizing flight time and energy consumption. Trajectory planning is typically divided into two approaches: one-stage and two-stage. The one-stage approach directly produces a feasible trajectory from the current state to the goal, while the two-stage approach uses a front-end to find a collision-free path or safe flight corridor and a back-end to optimize the trajectory.
Over the years, UAV planning techniques have matured significantly. In the following sections, we will review existing trajectory planning methods for both vision-based and LiDAR-based platforms and explore their application to LiDAR-equipped UAVs.

\subsubsection{One-stage approaches}
One-stage approaches, illustrated in Fig.\ref{fig:planning}A, evolved from traditional path-search methods and have been extensively applied in both ground and aerial vehicles. In 1999, LaValle \textit{et al.}\cite{lavalle1999randomized} introduced the Kinodynamic RRT (RRT*) algorithm, later extended in~\cite{webb2013kinodynamic, li2016asymptotically}. These methods replace the steering function in RRT* with a dynamically feasible one by solving optimal control (OC) or boundary value problem (BVP) formulations, ensuring the generated path respects the vehicle's dynamics.
In 2009, Likhachev \textit{et al.}\cite{likhachev2009planning} proposed a Multi-resolution Lattice State Space approach, discretizing the configuration space into states connected by dynamically feasible paths. MacAllister \textit{et al.}\cite{macallister2013path} applied this method to UAVs in 2013 using the AD* algorithm to generate smooth, feasible trajectories. Pivtoraiko \textit{et al.}~\cite{pivtoraiko2013incremental} further enhanced UAV navigation by integrating a motion primitive library into an online replanning framework for unknown environments.
For LiDAR-based UAVs, Zhang \textit{et al.}\cite{zhang2018pcap, zhang2019pcal, zhang2020falco} utilized motion primitive-based methods to achieve high-speed flight (over 10 m/s) in unknown environments. In contrast, Liu \textit{et al.}\cite{liu2017planning} introduced a trajectory search method based on the Linear Quadratic Minimum Time (LQMT) problem and extended it by considering the UAV’s shape for whole-body motion planning, enabling precise maneuvers through narrow gaps~\cite{liu2017search}.

\subsubsection{Two-stage approaches}
Compared to one-stage approaches, two-stage approaches—comprising front-end and back-end modules—have gained traction in UAV planning (Fig.~\ref{fig:planning}B). The front-end generates a collision-free path or safe flight corridor without considering dynamic feasibility, while the back-end optimizes a dynamically feasible trajectory to minimize factors such as time or energy. {We briefly divided the two-stage approaches into three categories: waypoint-based, distance field-based, and safe flight corridor (SFC)-based methods.}

    {\paragraph{{Waypoint-based:}}~The waypoint-based approach consists of a frontend module that selects a set of collision-free waypoints and a backend module that optimizes a feasible trajectory to sequentially pass through these waypoints.} Mellinger \textit{et al.}\cite{mellinger2011minimum} achieved UAV navigation by optimizing the trajectory using Quadratic Programming (QP) to minimize energy consumption (snap) while ensuring the UAV passes through all waypoints. {However, since these waypoints are user-defined, this method struggles in unknown environments.} Richter \textit{et al.}\cite{richter2016polynomial} extended this by proposing a closed-form solution for minimum snap trajectory optimization,
using RRT*~\cite{lavalle1998rapidly} {in the frontend to select waypoints at fixed intervals, followed by trajectory refinement. Similarly, Ji et al. \cite{ji2021mapless} proposed a mapless frontend that iteratively selects waypoints within a spherical free space, paired with alternating minimization \cite{wang2020alternating} for optimization.}
However, deviations between the path and optimized trajectory often require iterative waypoint insertion {to ensure a collision-free result, which compromises computational efficiency and trajectory quality.}

    {\paragraph{{Distance Field-based:}}~To address the limitation of waypoint-based methods and offer enhanced freedom in trajectory optimization,} Ding \textit{et al.}\cite{ding2019efficient} introduced elastic optimization to enforce collision-free constraints during optimization. Other methods use Euclidean Signed Distance Fields (ESDFs)\cite{zucker2013chomp, oleynikova2016continuous, zhou2021raptor} to maintain collision-free paths. However, building and updating the ESDF-map can be {can be computationally heavy.} Zhou \textit{et al.}~\cite{zhou2020ego} proposed a gradient-based method that eliminates ESDFs, offering a more efficient solution while maintaining dynamic feasibility.

{\paragraph{{Safe Flight Corridor-based:}}~The SFC-based approaches have also been widely studied. These methods use convex 3D shapes to represent collision-free corridors, constraining trajectories to ensure safety. Spheres~\cite{gao2019flying, ren2022bubble}, axis-aligned cubes~\cite{chen2015real}, and polyhedra~{\cite{liu2017planning, tordesillas2021faster,chen2023flying,ren2023online}} are common representations. Gao \textit{et al.} generated spherical SFCs using RRT*, while Ren \textit{et al.}\cite{ren2022bubble} used sample-based methods to generate them around a collision-free path. Chen \textit{et al.}\cite{chen2015real} developed axis-aligned cubes using octree structures, and Deits \textit{et al.}\cite{deits2015computing} proposed Iterative Region Inflation by Semidefinite Programming (IRIS) to generate large polyhedra with Semi-Definite Programming. Liu \textit{et al.}\cite{liu2017planning} introduced the more efficient Region Inflation with Line Search (RILS),
and Wang \textit{et al.}\cite{wang2025fast} developed the Fast Iterative Regional Inflation (FIRI) algorithm for high-quality polyhedra.

Although sphere- and cube-based SFCs are easy to generate, they may underrepresent free space, leading to suboptimal trajectories. Polyhedron-based SFCs better approximate free space but are computationally more expensive. The back-end optimization typically formulates SFCs as constraints to ensure collision-free trajectories. Chen \textit{et al.}\cite{chen2015real} and Liu \textit{et al.}\cite{liu2017planning} solved the optimization using QP, Gao \textit{et al.}\cite{gao2019flying} applied SOCP, and Tordesillas \textit{et al.}\cite{ tordesillas2021faster} used MIQP for trajectory optimization.
In 2022, Wang \textit{et al.}~\cite{wang2022geometry} introduced MINCO, a spatial-temporal deformation method that optimizes both spatial and temporal aspects of trajectories efficiently. The problem is reformulated as an unconstrained nonlinear optimization and solved using the quasi-Newton method.

\subsubsection{Discussion}

The one-stage and two-stage approaches each have distinct advantages and limitations. One-stage approaches are more efficient for short planning horizons but require simplifications to manage computational complexity, often leading to suboptimal solutions. As the planning horizon increases, their time consumption also rises. In contrast, two-stage approaches generate more accurate and dynamically feasible trajectories in less time. The front-end efficiently finds a collision-free path or flight corridor without considering UAV dynamics, while the back-end optimizes the trajectory within a smaller solution space, reducing complexity and time. Moreover, the two-stage approach can incorporate various constraints, such as energy efficiency, time minimization, and safety, making it highly adaptable.

For LiDAR-based UAVs, the challenge lies in leveraging LiDAR's long-range, high-rate 3D measurements to enhance navigation and control. Effective planning must capitalize on LiDAR's long sensing range to enable obstacle avoidance and high-speed flight. Two-stage methods are better suited for this, and we explored three types: 1) Waypoint-based, 2) Distance field-based, and 3) Safe Flight Corridor (SFC)-based methods.

Waypoint-based methods are simple but rely on heuristic waypoint placement, which can yield suboptimal results. Distance field-based methods, such as those using ESDF maps, offer smooth gradients for optimization but are computationally expensive, with non-convex costs prone to local minima. In contrast, corridor-based methods enforce collision-free constraints through convex-shaped SFCs, ensuring efficient, high-quality optimization. With advances in differentiable optimization, such as MINCO~\cite{wang2022geometry}, SFC-based methods have become ideal for LiDAR-based UAVs. For example, Ren \textit{et al.}\cite{ren2022bubble} demonstrated a LiDAR-based UAV navigating cluttered environments at over 13.7 m/s using spherical SFCs and MINCO-based optimization. Additionally, corridor-based methods integrate seamlessly with MPC backends\cite{liu2023integrated}, enabling low-latency navigation and real-time obstacle avoidance for sudden threats.

\subsection{Control}
Once a collision-free and dynamically feasible trajectory is generated, the control module is responsible for executing the commands from the planning process. Trajectory tracking control methods have been extensively studied over the past decades. Thanks to the high-accuracy and low-latency state estimation enabled by LiDAR sensors, trajectory tracking control can achieve the precision necessary for most applications.

The most commonly used control method for UAVs is the Proportional-Integral-Derivative (PID) controller~\cite{salih2010flight, li2011dynamic}. The PX4 Autopilot software~\cite{px4ctrl}, widely adopted in UAV systems, employs a cascaded control architecture that combines P and PID controllers.
This method is straightforward, easy to tune, and extremely efficient, making it ideal for running on embedded systems.

Beyond PID, the Linear Quadratic Regulator (LQR)\cite{reyes2013lqr, kuantama2018feedback} is another widely used optimal control method. LQR operates by minimizing a specific cost function, typically providing better performance than PID. Model Predictive Control (MPC) is also commonly used in UAV control due to its ability to handle system constraints and uncertainties, making it especially effective in the presence of disturbances. Hanover {\textit{et al.}}\cite{hanover2021performance} proposed an adaptive nonlinear MPC for high-accuracy trajectory tracking during high-speed flights, while Fang {\textit{et al.}}\cite{nan2022nonlinear} developed a nonlinear MPC followed by an Incremental Nonlinear Dynamic Inversion (INDI) block to enhance robustness. Sun {\textit{et al.}}\cite{sun2022comparative} incorporated aerodynamic forces into a nonlinear MPC (NMPC) and INDI-based control framework for quadrotors, and Lu {\textit{et al.}}~\cite{mpc_luguozheng} introduced a singularity-free, on-manifold MPC framework with minimal parameterization for trajectory tracking. In this approach, the system state is mapped to local coordinates around each reference point on the trajectory, transforming the manifold-constrained control problem into a standard quadratic programming (QP) form that can be solved efficiently.

With advances in onboard computing and numerical optimization techniques, MPC-based methods are becoming increasingly popular for UAV control, offering superior tracking performance while being computationally feasible in real time.

\subsection{Integrated planning and control}

Beyond classical planning and control methods, integrated planning and control approaches have also been developed. Lindqvist {\textit{et al.}}\cite{lindqvist2020nonlinear} modeled obstacles as spheres and proposed a nonlinear model predictive control (NMPC)-based method for obstacle avoidance.
    {Similarly, Soria \textit{et al.} \cite{soria2021predictive} incorporated a quadratic penalty term into an NMPC framework, while Ali \textit{et al.} \cite{ali2024mpc} proposed a control barrier function (CBF)-based MPC approach to integrate planning and control {to} achieve obstacle avoidance. However, both methods assume that obstacle locations are known, with each obstacle point (e.g., a LiDAR point) imposing a single constraint. This assumption leads to an excessive computational burden, particularly with dense LiDAR data, and limits their applicability in unknown environments.}
    {Santos {\textit{et al.}}\cite{santos2017novel} proposed a novel null-space-based trajectory tracking controller with collision avoidance for UAVs.}
Liu {\textit{et al.}}\cite{liu2023integrated,liu2025slope} introduced Integrated Planning and Control (IPC), which incorporates polyhedron-shaped SFC constraints into an MPC framework, enabling collision-aware real-time control.
In this approach, the front-end generates polyhedron-shaped SFCs based on LiDAR point cloud data, while the back-end directly generates control commands using an MPC-based framework. The IPC method is efficient, accurate, and robust, making it particularly suitable for LiDAR-based UAVs.

\renewcommand{\arraystretch}{1.2} 

\begin{table*}
    \begin{threeparttable}[ht]
        \vspace{-0.3cm}
        \fontsize{6.5}{6.0}\selectfont
        \centering
        \caption{Open Source Resources for LiDAR-based Autonomous UAVs}
        \label{tab:open}
        \begin{tabular}{m{3cm}m{4cm}m{1.0cm}m{8.4cm}}
            \midrule
            \textbf{Name and Reference}                       & \textbf{Category}                    & \textbf{Year} & \textbf{Link}                                                                     \\
            \midrule
            \arrayrulecolor{gray} 
            OmniNxt ~\cite{liu2024omninxt}                    & Hardware \& Software Stack (Vision)  & 2024          & \url{https://github.com/HKUST-Aerial-Robotics/OmniNxt}                            \\
            UniQuad ~\cite{zhang2024uniquad}                  & Hardware  (LiDAR \& Vision)          & 2024          & \url{https://github.com/HKUST-Aerial-Robotics/UniQuad}                            \\
            Agilicious ~\cite{foehn2022agilicious}            & Hardware \& Software Stack  (Vision) & 2023          & \url{https://github.com/uzh-rpg/agilicious}                                       \\
            PULSAR ~\cite{chen2023self}                       & Hardware  (LiDAR)                    & 2023          & \url{https://github.com/hku-mars/PULSAR}                                          \\
            Fast-Drone-250                                    & Hardware\& Software Stack  (Vision)  & 2022          & \url{https://github.com/ZJU-FAST-Lab/Fast-Drone-250}                              \\

            \specialrule{0.1pt}{0pt}{0pt} 
            Aerial Gym ~\cite{kulkarni2023aerial}             & Simulator                            & 2024          & \url{https://github.com/ntnu-arl/aerial_gym_simulator}                            \\
            MARSIM ~\cite{kong2023marsim}                     & Simulator                            & 2023          & \url{https://github.com/hku-mars/MARSIM}                                          \\
            flightmare ~\cite{song2021flightmare}             & Simulator                            & 2021          & \url{https://github.com/uzh-rpg/flightmare}                                       \\
            AirSim ~\cite{madaan2020airsim}                   & Simulator                            & 2020          & \url{https://microsoft.github.io/AirSim/}                                         \\
            rotorS ~\cite{furrer2016rotors}                   & Simulator                            & 2016          & \url{https://github.com/ethz-asl/rotors_simulator}                                \\
            \specialrule{0.1pt}{0pt}{0pt} 
            CLEARLAB ~\cite{xu2024heuristic}                  & Software Stack                       & 2024          & \url{https://github.com/Zhefan-Xu/CERLAB-UAV-Autonomy}                            \\
            CMU ~\cite{zhang2020falco}                        & Software Stack                       & 2020          & \url{https://github.com/HongbiaoZ/autonomous_exploration_development_environment} \\
            \specialrule{0.1pt}{0pt}{0pt} 
            DVLC ~\cite{koide2023general}                     & Calibration                          & 2023          & \url{https://github.com/koide3/direct_visual_lidar_calibration}                   \\
            LI-Init ~\cite{zhu2022robust}                     & Calibration                          & 2022          & \url{https://github.com/hku-mars/LiDAR_IMU_Init}                                  \\
            LVI-ExC ~\cite{wang2022lvi}                       & Calibration                          & 2022          & \url{https://github.com/peterWon/LVI-ExC}                                         \\
            lidar-camera-calib ~\cite{yuan2021pixel}          & Calibration                          & 2021          & \url{https://github.com/hku-mars/livox_camera_calib}                              \\
            Kalibr ~\cite{rehder2016extending}                & Calibration                          & 2016          & \url{https://github.com/ethz-asl/kalibr}                                          \\
            \specialrule{0.1pt}{0pt}{0pt} 
            FAST-LIVO2 ~\cite{zheng2024fast}                  & Localization                         & 2024          & \url{https://github.com/hku-mars/FAST-LIVO2}                                      \\
            Swarm-LIO2 ~\cite{zhu2024swarm}                   & Localization                         & 2024          & \url{https://github.com/hku-mars/Swarm-LIO2}                                      \\
            Point-LIO ~\cite{he2023point}                     & Localization                         & 2023          & \url{https://github.com/hku-mars/Point-LIO}                                       \\
            FAST-LIO2 ~\cite{xu2022fast}                      & Localization                         & 2022          & \url{https://github.com/hku-mars/FAST_LIO}                                        \\
            LIO-SAM ~\cite{shan2020lio}                       & Localization                         & 2020          & \url{https://github.com/TixiaoShan/LIO-SAM}                                       \\
            A-LOAM                                            & Localization                         & 2020          & \url{https://github.com/HKUST-Aerial-Robotics/A-LOAM}                             \\
            \specialrule{0.1pt}{0pt}{0pt} 
            ROG-Map ~\cite{ren2024rog}                        & Mapping                              & 2024          & \url{https://github.com/hku-mars/ROG-Map}                                         \\
            D-Map ~\cite{cai2023occupancy}                    & Mapping                              & 2024          & \url{https://github.com/hku-mars/D-Map}                                           \\
            FIESTA ~\cite{han2019fiesta}                      & Mapping                              & 2019          & \url{https://github.com/HKUST-Aerial-Robotics/FIESTA}                             \\
            Voxblox ~\cite{oleynikova2017voxblox}             & Mapping                              & 2017          & \url{https://github.com/ethz-asl/oleynikova2017voxblox}                           \\
            OctoMap ~\cite{hornung2013octomap}                & Mapping                              & 2013          & \url{https://github.com/OctoMap/octomap}                                          \\
            \specialrule{0.1pt}{0pt}{0pt} 
            {SUPER ~\cite{ren2025safety}   }                  & Planner                              & 2025          & \url{https://github.com/hku-mars/SUPER}                                           \\
            PMM ~\cite{teissing2024pmm}                       & Planner                              & 2024          & \url{https://github.com/ctu-mrs/pmm_uav_planner}                                  \\
            GCOPTER ~\cite{wang2022geometry}                  & Planner                              & 2022          & \url{https://github.com/ZJU-FAST-Lab/GCOPTER}                                     \\
            mintime-replan ~\cite{penicka2022minimum}         & Planner                              & 2022          & \url{https://github.com/uzh-rpg/sb_min_time_quadrotor_planning}                   \\
            FASTER ~\cite{tordesillas2021faster}              & Planner                              & 2021          & \url{https://github.com/mit-acl/faster}                                           \\
            time optimal ~\cite{foehn2021time}                & Planner                              & 2021          & \url{https://github.com/uzh-rpg/rpg_time_optimal}                                 \\
            TRR ~\cite{gao2020teach}                          & Planner                              & 2020          & \url{https://github.com/HKUST-Aerial-Robotics/Teach-Repeat-Replan}                \\
            FastPlanner ~\cite{zhou2021raptor}                & Planner                              & 2020          & \url{https://github.com/HKUST-Aerial-Robotics/Fast-Planner}                       \\
            EGO-Planner ~\cite{zhou2020ego}                   & Planner                              & 2020          & \url{https://github.com/ZJU-FAST-Lab/ego-planner}                                 \\
            \specialrule{0.1pt}{0pt}{0pt} 
            IPC ~\cite{liu2023integrated}                     & Planner \& Controller                & 2023          & \url{https://github.com/hku-mars/IPC}                                             \\
            Geometry Control                                  & Controller                           & 2021          & \url{https://github.com/yorgoon/minimum-snap-geometric-control}                   \\
            PMPC ~\cite{Falanga2018Perception}                & Controller                           & 2018          & \url{https://github.com/uzh-rpg/rpg_mpc}                                          \\
            RPG Quad Control ~\cite{faessler2017differential} & Controller                           & 2018          & \url{https://github.com/uzh-rpg/rpg_quadrotor_control}                            \\
            QuadRotor-Control                                 & Controller                           & 2018          & \url{https://github.com/srikantrao/QuadRotor-Control}                             \\
            \specialrule{0.1pt}{0pt}{0pt} 
            PX4                                               & Autopilot                            & -             & \url{https://github.com/PX4}                                                      \\
            Betaflight                                        & Autopilot                            & -             & \url{https://github.com/betaflight/betaflight}                                    \\
            ArduPilot                                         & Autopilot                            & -             & \url{https://github.com/ArduPilot/ardupilot}                                      \\
            \arrayrulecolor{black} 
            \midrule
        \end{tabular}
    \end{threeparttable}%
    \vspace{-0.5cm}
\end{table*}

\section{Open Source Projects for LiDAR-based Autonomous UAVs}
\label{sec:open}
Building an autonomous LiDAR-based system from hardware to software is a complex, system-wide task. Open-source resources provide a valuable foundation for researchers and developers. In TABLE~\ref{tab:open}, we present a curated list of open-source resources relevant to LiDAR-based autonomous UAVs, including hardware, software stacks, simulators, calibration tools, localization, mapping, planning, control, and autopilot systems. Part of the table is adapted from~\cite{hanover2024autonomous}, with additional recent resources specifically focused on LiDAR-based UAVs.
A typical pipeline for building an autonomous UAV involves:

\begin{enumerate}
    \item Designing the hardware structure, incorporating essential sensors, computational units, and actuators.
    \item 	Calibrating sensors such as LiDAR and IMUs, including both spatial and temporal alignment.

    \item Deploying navigation software, covering perception, planning, and control modules.
    \item Testing the software stack in simulation to evaluate performance.

    \item Integrating the software with hardware and conducting real-world tests.
\end{enumerate}

{The open-source resources in Table~\ref{tab:open} cover all stages, forming a solid framework for LiDAR-based autonomous UAVs. Check our GitHub repo at \url{https://github.com/hku-mars/LiDAR-UAV-Autonomy} for updates.}


\section{Applications}
\label{sec:app}

\subsection{Industrial Applications}
Autonomous LiDAR-equipped UAVs have been used in many industrial applications, including inspection \cite{car2020autonomous, castelar2024lidar, bolourian2017high, bolourian2020lidar, guan2021uav,paneque2022autonomous}, agriculture \cite{car2020autonomous,castelar2024lidar,bolourian2017high, bolourian2020lidar,guan2021uav,paneque2022autonomous}, search {and} rescue \cite{tian2020search, doshida2021evaluation}, and package delivery \cite{dissanayaka2022visual}.

\subsubsection{Inspection}
Inspection tasks can be categorized into areas such as wind-turbine blades~\cite{car2020autonomous, castelar2024lidar}, bridges~\cite{bolourian2017high, bolourian2020lidar}, and power lines~\cite{guan2021uav, paneque2022autonomous}. {As shown in Fig. \ref{fig:uav_platform}A}, LiDAR sensors serve as primary tools for measuring geometric data and creating accurate 3D models of structures for analysis. Additionally, LiDAR can support inspections where cameras struggle, such as detecting thin power lines. Accurate LiDAR measurements combined with point cloud classification techniques can aid UAVs in power line-following tasks~\cite{wang2019hierarchical}.

\subsubsection{Agriculture}
Thanks to the accurate measurement of LiDAR, the users can pinpoint structures or zones of interest and highlight surface degradation or vegetation growth, which makes the use of LiDAR popular in the application of precision agriculture \cite{rivera2023lidar}. By processing and utilizing the data acquired by the LiDAR-equipped UAV, many tasks like the estimation of yield \cite{xu2020estimation}, health monitoring \cite{gomes2020geltip,caras2024monitoring}, height monitoring \cite{liu2020analysis, zhang2021high}, or tree detecting \& digitization \cite{itakura2018automatic, moreno2020ground} can be done automatically, leading to a smarter and precise agriculture.

\subsubsection{Search {and} rescue \& package delivery}
Currently, most UAVs used for search and rescue and package delivery rely on vision-based systems. However, LiDAR sensors show an advantage because they are not affected by lighting conditions and can provide precise measurements even during high-speed flights. This capability enables UAVs to perform essential functions such as obstacle avoidance, environmental mapping, and autonomous navigation even in crowded or low-light environments, which are important for successful search and rescue as well as efficient package delivery. So far, the LiDAR-equipped UAVs are already used in search {and} rescue \cite{tian2020search, doshida2021evaluation} and package delivery \cite{dissanayaka2022visual}. With the performance improvement and cost reduction of LiDAR sensors, more LiDAR-equipped UAVs will be used in these applications.

\subsection{Autonomous Aerial Platforms}
{
    Leveraging LiDAR’s strengths has driven major advancements in navigation, environmental perception, and flight speed. These advancements have unlocked new aerial platforms that were not achieved before.  This section examines two notable types of autonomous aerial platforms—self-rotating UAVs and vertical take-off and landing (VTOL) UAVs—that utilize LiDAR technology to achieve higher levels of autonomy.}

\subsubsection{Self-rotating UAV}
Self-rotating UAVs have a continuous body rotation motion during flight. This characteristic can obviously extend the limited FoV of the visual sensors (e.g., camera, LiDAR). Most of the existing UAVs {\cite{piccoli2017piccolissimo,zhang2016controllable,win2021design,bai2022bioinspired,bai2022splitflyer}} cannot achieve autonomous navigation since they have insufficient payload capability to carry sensor for environment observation as well as onboard computer for online computation. One self-rotating UAV \cite{jameson2012lockheed} equipped a camera for attitude estimation but it cannot perform autonomous navigation because it has no position and velocity estimation and no mapping ability. Another UAV that can estimate the full states during self-rotating uses down-facing event camera \cite{sun2021autonomous}. Down-facing camera ease the difficulty of state estimation but it is not beneficial for autonomous navigation since its FoV is limited downward and also cannot be extended by the self-rotation.

\begin{figure}[htbp]
    \centering
    \includegraphics[width=0.48\textwidth]{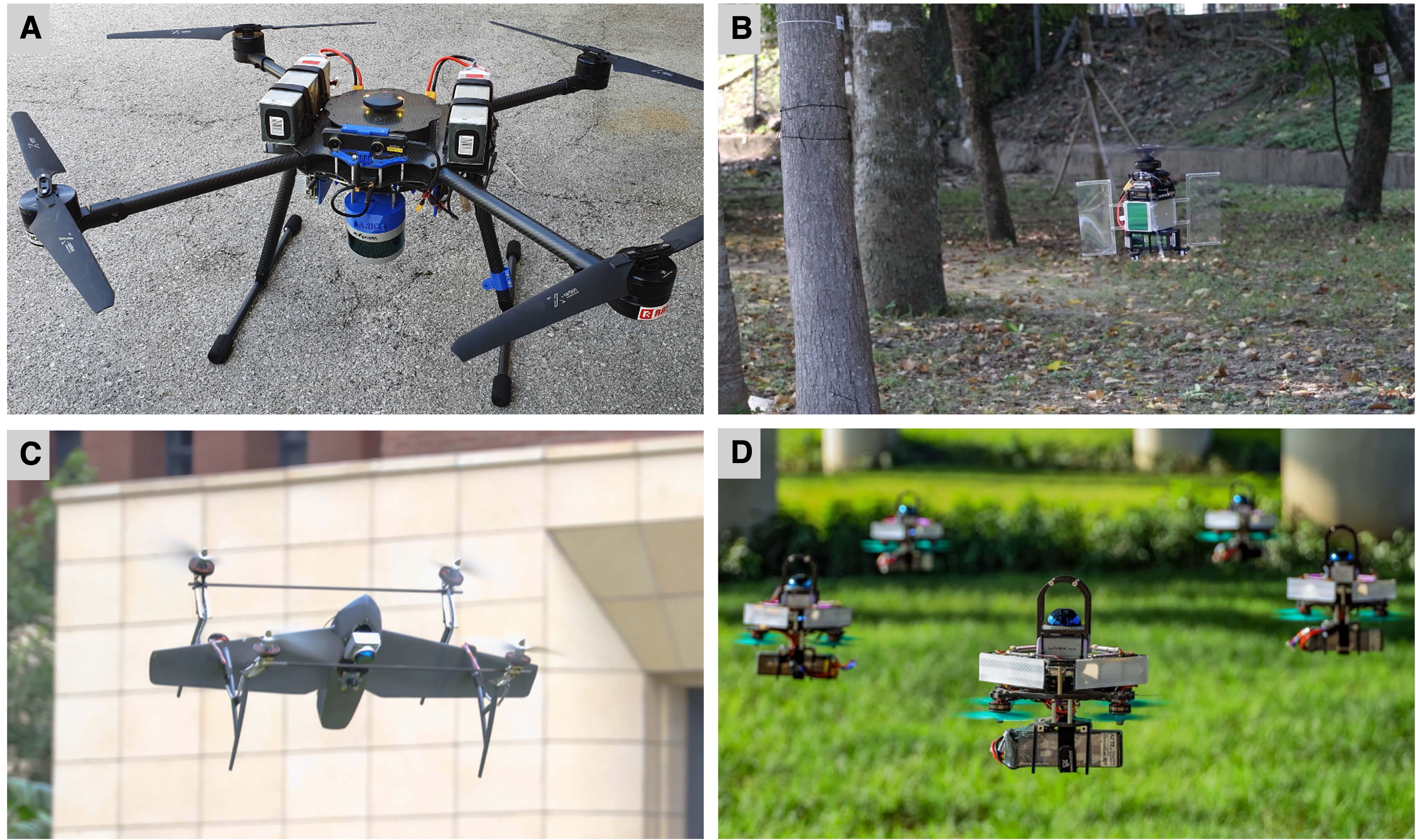}
    \caption{{LiDAR's application on different UAV platforms. (\textbf{A}) A quad-rotor UAV for wind-turbine blade inspection \cite{car2020autonomous}. (\textbf{B}) An single-actuator UAV achieving autonomous flight with extended LiDAR's FoV via self-rotation \cite{chen2023self}. (\textbf{C}) A VTOL UAV for autonomous flight in GNSS-denied and cluttered environments \cite{lu2025autonomous}. (\textbf{D}) UAV swarm system. \cite{zhu2024swarm}.}
    }
    \label{fig:uav_platform}
    \vspace{-0.5cm}
\end{figure}

The only one exception is a LiDAR-based self-rotating UAV named PULSAR \cite{chen2023self} {as shown in Fig. \ref{fig:uav_platform}B}. It achieved compete autonomous flight in an unknown GNSS-denied environment by only using the LiDAR sensor. At the same time, the limit conical FoV of the LiDAR sensor is extended through self-rotation to cover most area around the UAV, which obviously improved the mapping efficiency and obstacle perception ability. Unlike the camera may have image motion blur caused by fast self-rotation resulting in harder state estimation and mapping, the LiDAR sensor can realize accurate state estimation and mapping during high-speed self-rotation. For each LiDAR point measurement, the time interval is very short such that the self-rotation of the UAV can be neglected, and hence the LiDAR sensor is more suitable for autonomous self-rotating UAVs.

\subsubsection{Vertical Take-off and Landing UAVs}
Vertical takeoff and landing (VTOL) UAVs can take off and land vertically like typical multi-rotor UAVs and achieve long-range and high-speed flight with high efficiency similar to fixed-wing UAVs at the same time. Since their aerodynamics are highly nonlinear with large variation of angle of attack during transition and high flight speed, achieving the autonomy of VTOL UAVs is a very challenging task. A LiDAR-based VTOL UAV platform is proposed in \cite{chen2023swashplateless} for potential terrain-following flight, but the autonomous flight is not achieved so far. In \cite{lu2024trajectory}, the differential flatness of tail-sitter VTOL UAV with quadrotor configuration is proofed and verified, which is a huge advance to realize high-speed autonomy of this type of UAV. The first tail-sitter VTOL UAV that achieves fully autonomous flight ability is reported in \cite{lu2025autonomous} {as shown in Fig. \ref{fig:uav_platform}C}. It equips a LiDAR sensor to estimate states and to build environmental map as well as uses an efficient feasibility-first solver to optimize flight trajectory within complex aerodynamic constrains during a short time interval. As a result, this LiDAR-based VTOL UAV achieved 15-m/s high-speed flight in clustered environments such as underground parking lot and ourdoor park, fully autonomously.

\subsection{UAV Swarm Applications}
Multi-UAV systems offer significant advantages over single-UAV systems, including enhanced fault tolerance, high scalability, and increased coverage and efficiency. The failure of one UAV does not compromise the mission, as others can continue operating, ensuring reliability and mission success. Additionally, these systems enable comprehensive data collection from multiple perspectives, facilitate cooperative behaviors for improved task execution, and can reduce operational costs despite a higher initial investment. Overall, the robustness and flexibility of multi-UAV systems make them essential for various complex applications in the evolving landscape of autonomous aerial operations, such as swarm aerial tracking and swarm exploration.

\subsubsection{Swarm Aerial Tracking} Swarm tracking empowers UAVs to autonomously identify, follow, and monitor specific targets in real time. Camera-based multi-UAV tracking has drawn increasing attention in the recent literature\cite{zhou2022swarm, tallamraju2019active, bucker2021you, ho20213d}, which fully leverages the advancement of camera-based detection. Compared to cameras, LiDAR sensors can provide more accurate measurements on environments and targets . In 2020, Bonatti. \textit{et al.} \cite{bonatti2020autonomous} presented an aerial tracking UAV which mounted a VLP-16 LiDAR for environmental sensing. However, the target in \cite{bonatti2020autonomous} is still measured with a camera, and the system only contained a single UAV. In 2023, Yin. \textit{et al.} \cite{yin2023decentralized} proposed a swarm tracking system that solely utilizes Mid360 LiDARs as the sensors. {In 2024, Zhu. \textit{et al.} \cite{zhu2024swarm} achieved dynamic formation and human object tracking for UAV swarm system (as shown on Fig. \ref{fig:uav_platform}D) based on LiDAR sensors.} The proposed system achieved visibility-aware tracking upon a high-reflective-tape-marked drone as the target in cluttered scenes, which validated the practicability of LiDAR-based cooperative tracking.

\subsubsection{Swarm Exploration} Swarm exploration refers to the coordinated use of multiple UAVs to autonomously explore and gather information about an environment. This approach leverages the strengths of swarm robotics and distributed systems to achieve simultaneous data collection and hence higher efficiency than single-UAV exploration. Compared to the camera-based swarm exploration \cite{zhou2023racer}, LiDAR-based ones can provide superior sensing ranges and discover the detailed geometry of the environments, which helps render more accurate frontiers for exploration \cite{dong2024fast}. Yu. \textit{et al.} \cite{yu2021smmr} proposed a multi-robot multi-target potential field model to select the best frontier goal and validated the method physically with two ground vehicles equipped with 2D-LiDARs. However, this work only considered 2D spaces. In \cite{dong2024fast}, a voronoi partition strategy was presented for task allocation and a real-world swarm of five Mid360 UAVs was deployed to validate the proposed method.

\begin{figure*}[htb]
    \centering
    \includegraphics[width=0.9\textwidth]{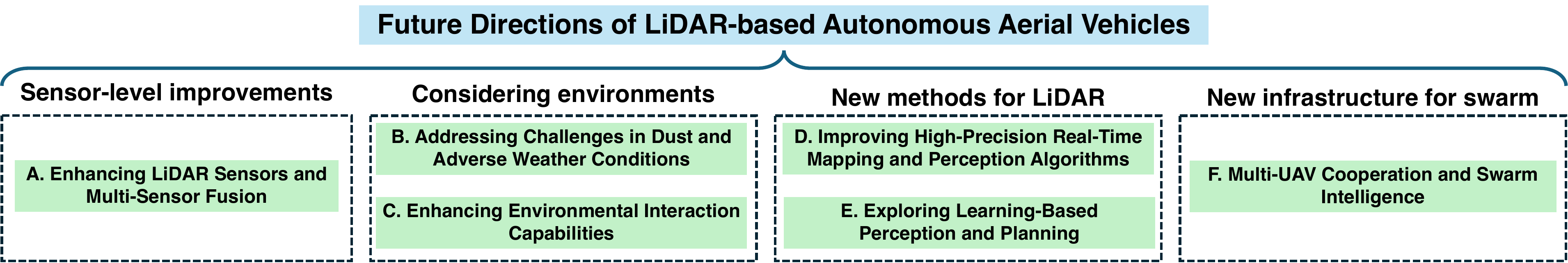}
    \caption{Summary and classification of future directions.
    }
    \label{fig:future_dir}
    \vspace{-0.45cm}
\end{figure*}

\section{Discussion and Future Directions}
\label{sec:disc}
The rapid advancement of LiDAR has significantly enhanced UAV's autonomy, however, there are still considerable gaps to perfection. The following sections outline key challenges and future directions for enhancing the autonomy, perception, and performance of LiDAR-equipped UAVs through sensor advancements and algorithmic innovations that could elevate their capabilities to the next level{, which are summarized and classified in Fig. \ref{fig:future_dir}.}

\vspace{-0.3cm}

\subsection{Enhancing LiDAR Sensors and Multi-Sensor Fusion}

Current LiDAR sensors face limitations related to weight, cost, and difficulties in detecting certain surfaces, such as {mirrors~\cite{yang2010solving}}, glass and water, which impact UAV performance. Future research should prioritize the development of advanced LiDAR technologies, such as MEMS-based and METASURFACE LiDAR~\cite{mems_lidar, metasurface}, to reduce weight and enhance the payload capacity of UAVs.

An emerging area of interest is the development of pure solid-state LiDARs, such as flash LiDAR. Flash LiDAR offers significant potential due to its compact size, reduced weight, and simplified design with no moving parts. This technology allows for further miniaturization, which is particularly beneficial for UAVs where weight and space are critical constraints. {Improvements in detection range, resolution, and point cloud density could significantly enhance the precision and long-range sensing capabilities of UAVs equipped with flash LiDAR.}

    {Additionally, recent progress in hybrid sensor systems, such as long-range, high-density RGB-D systems integrating LiDAR and cameras, presents exciting opportunities to overcome current limitations. By combining LiDAR’s precise geometric data with cameras’ rich semantic and textural information, these systems can improve performance in tasks like object detection, scene segmentation, and path planning. Developing effective fusion techniques for these hybrid sensors is a key area for future research.}

    {Besides camera, i}ntegrating LiDAR with other sensors (e.g., GPS, radar) in a unified perception framework can further improve robustness and sensing capabilities. Adaptive sensor fusion algorithms that dynamically adjust to environmental conditions can enhance UAV reliability, effectively addressing corner cases involving reflective or transparent surfaces.

\subsection{Addressing Challenges in Dust and Adverse Weather Conditions}

Dust, rain, and fog pose significant challenges for LiDAR-based UAVs, as these conditions can disrupt LiDAR measurements and degrade system performance. Future efforts should focus on developing adaptive filtering algorithms and signal processing techniques to mitigate noise and scattering effects caused by such adverse conditions. Integrating sensor fusion methods that combine LiDAR with radar and thermal imaging can further improve UAVs’ ability to perceive and operate reliably. Additionally, machine learning-based correction models and robust data pre-processing strategies may enhance LiDAR data accuracy, ensuring dependable navigation and obstacle detection under challenging weather.

\subsection{Enhancing Environmental Interaction Capabilities}

Currently, aerial robots primarily serve as passive observers, avoiding obstacles in their environment with limited capability for interaction to ensure flight safety.
Future research should explore enabling LiDAR-based UAVs to actively interact with and even change their surroundings to achieve specific navigation tasks. Integrating robotic manipulation technologies could allow UAVs to perform operations such as opening doors~\cite{opendoor} or moving obstacles. Additionally, bio-inspired designs and the incorporation of flexible sensors could further expand UAVs’ operational capabilities, enabling active environmental interaction beyond mere obstacle avoidance.

\subsection{Improving High-Precision Real-Time Mapping and Perception Algorithms}

The long detection range and {high point rates} of LiDAR sensors present challenges for real-time mapping and perception algorithms~\cite{cai2023occupancy}. Optimizing computational performance in mapping algorithms, such as occupancy grid mapping, is crucial to fully leverage LiDAR's capabilities. Future research should prioritize more efficient update methods and data compression techniques to enhance memory efficiency. Exploring continuous mapping~\cite{li2024gmmap} and decremental mapping~\cite{cai2023occupancy} could further boost real-time performance. Additionally, deep learning-based approaches, such as implicit mapping, could be employed to process extensive LiDAR data while preserving maximum information in real-time.



\subsection{Exploring Learning-Based Perception and Planning}
{
    LiDAR provides rich geometric data that significantly enhances geometric perception methods~\cite{xu2022fast}; however, the implicit semantic information within LiDAR point clouds remains underexplored. A promising research direction involves integrating geometric and semantic information from LiDAR data to advance UAV navigation. By combining spatial geometry with contextual cues—such as object classification and scene semantics—UAVs could achieve superior scene understanding and decision-making in complex environments. The sparse and non-uniform nature of LiDAR data, however, poses a significant challenge, necessitating innovative algorithms to effectively fuse these data types and unlock their combined potential for robust, adaptive navigation.
}

{Learning-based methods have shown considerable promise in enhancing UAV planning~\cite{loquercio2021learning}. By integrating LiDAR’s long-range, high-accuracy measurements with learning-based approaches, UAVs could achieve more agile and high-speed flight. These methods leverage LiDAR’s detailed spatial data to optimize planning and control strategies, improving adaptability in dynamic and complex environments. Nevertheless, the relative sparsity of LiDAR measurements compared to vision sensors can limit the performance of learning-based approaches. Future research should prioritize developing algorithms that address this sparsity, enabling learning-based methods to fully harness LiDAR’s potential for enhanced UAV performance.}


\subsection{Multi-UAV Cooperation and Swarm Intelligence}

Most research to date has focused on individual UAV autonomy, but significant potential exists for exploring multi-UAV cooperation in complex missions~\cite{zhou2023racer,zhu2023swarm,zhu2024swarm,yin2023decentralized}. Developing collaborative perception and decision-making frameworks will enhance the task planning and adaptability of UAV swarms. This can facilitate efficient mission execution in large-scale, challenging environments, leveraging LiDAR’s capabilities for comprehensive, cooperative operations.

\section*{Acknowledgment}

 {This project was supported in part by a donation from DJI and in part by the Hong Kong General Research Fund (GRF) under grant number 17204523.}

\bibliographystyle{IEEEtran}
\bibliography{paper}

\vspace{-30pt}
\begin{IEEEbiography}[{\includegraphics[width=1in,height=1.25in,clip,]{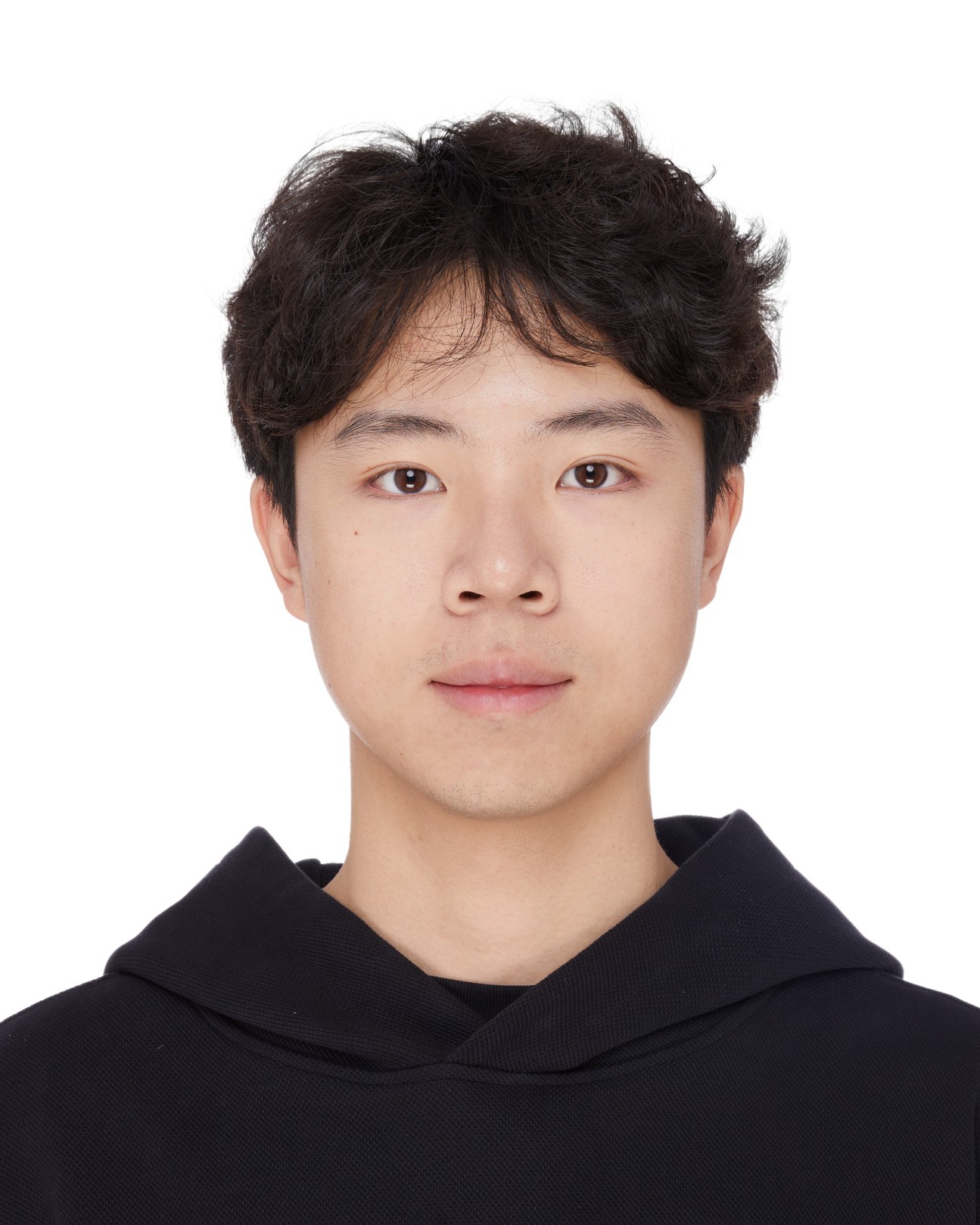}}]{Yunfan Ren}
    received his B.Eng. degree in Automation from Harbin Institute of Technology, Shenzhen, China, in 2020, and his Ph.D. degree in Robotics from The University of Hong Kong, HKSAR, China, in 2025. He is currently a Postdoctoral Researcher at the University of Zurich, Switzerland. His research focuses on autonomous systems, encompassing autonomous navigation, optimal control, reinforcement learning, and generative AI for robotic navigation.
\end{IEEEbiography}

\vspace{-30pt}
\begin{IEEEbiography}[{\includegraphics[width=1in,height=1.25in,clip,]{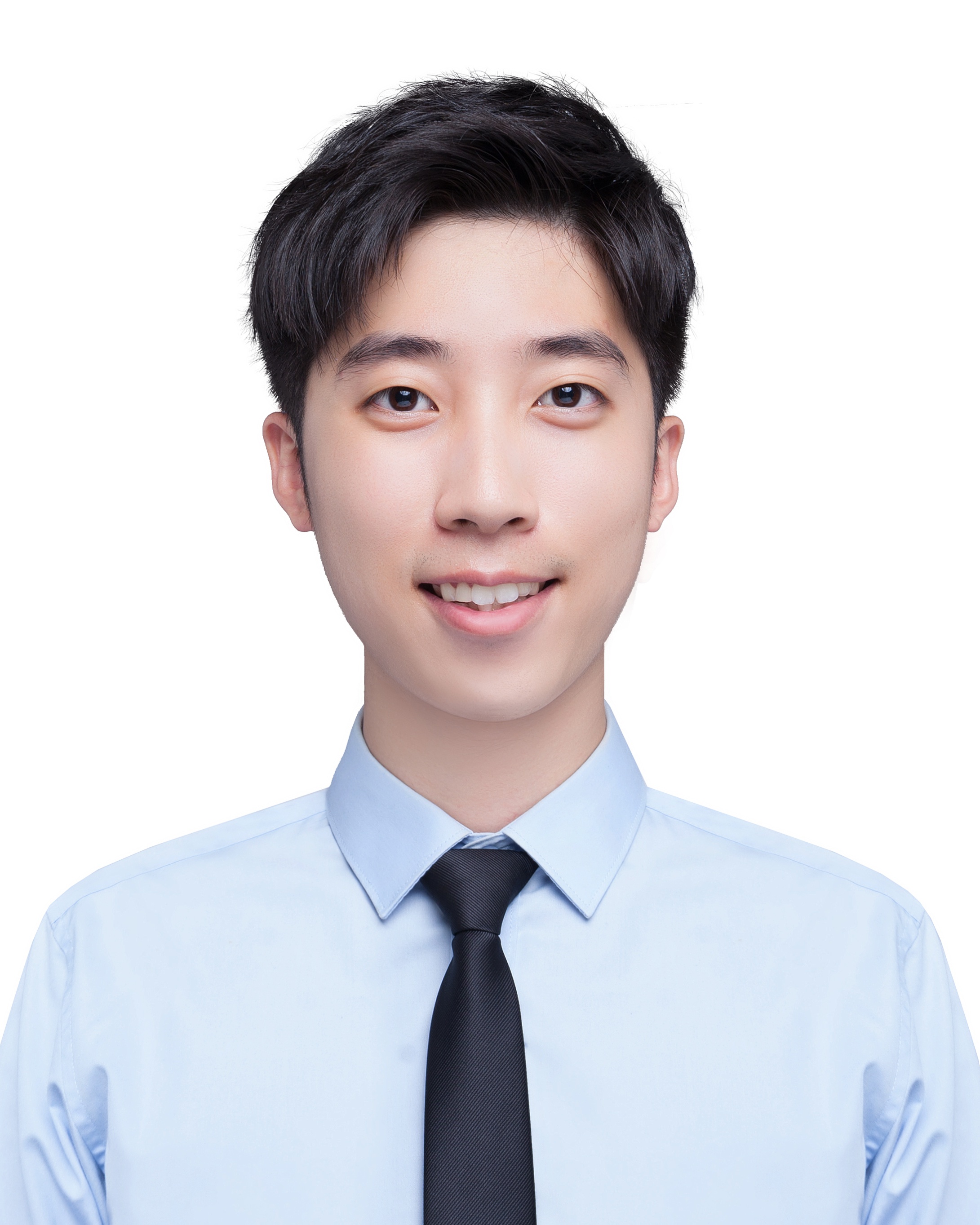}}]{Yixi Cai}
    received the B.Eng. degree in Automation, in 2020, from Beihang University (BUAA), Beijing, China, and the Ph.D. degree in Robotics, in 2024, from The University of Hong Kong, HKSAR, China.  He is currently a Postdoctoral Researcher at KTH, Stockholm, Sweden. His research interest lies in world representation for robotics, with a particular emphasis on efficient mapping techniques using light detection and ranging (LiDAR) for purposes such as localization, navigation, and exploration.
\end{IEEEbiography}

\vspace{-30pt}
\begin{IEEEbiography}[{\includegraphics[width=1in,height=1.25in,clip,]{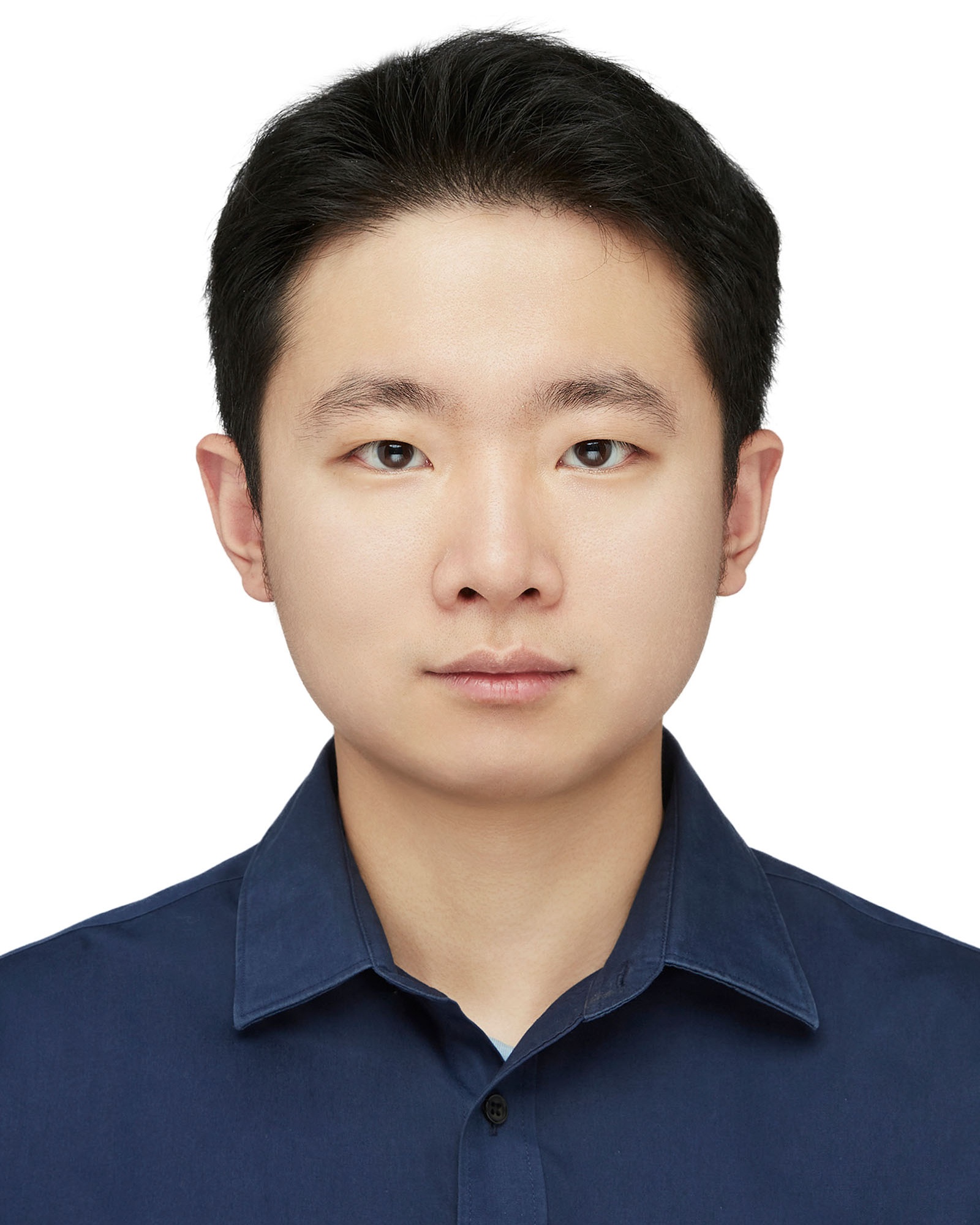}}]{Haotian Li}
    received a B.E. degree in Measuring and Control Technology and Instrumentations from Harbin Engineering University (HEU) in 2021. He is currently a Ph.D. candidate in the Department of Mechanical Engineering, the University of Hong Kong(HKU), Hong Kong, China. His current research interests are on robotics, with a focus on unmanned aerial vehicle (UAV) design, and sensor fusion.
\end{IEEEbiography}

\begin{IEEEbiography}[{\includegraphics[width=1in,height=1.25in,clip,]{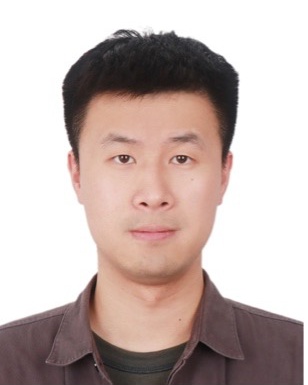}}]{Nan Chen} received the B.Eng. degree in automation and the M.Eng. degree in control theory and control engineering, both from Shenzhen University, Shenzhen, China. Then, he received the Ph.D. degree in Robotics from the Department of Mechanical Engineering, The University of Hong Kong, Hong Kong, China. He is currently a Research Assistant Professor with the Department of Mechanical Engineering, The University of Hong Kong, Hong Kong, China. His research interests include design, control, planning, and navigation of unmanned aerial vehicles.
\end{IEEEbiography}

\vspace{-60pt}
\begin{IEEEbiography}[{\includegraphics[width=1in,height=1.25in,clip,]{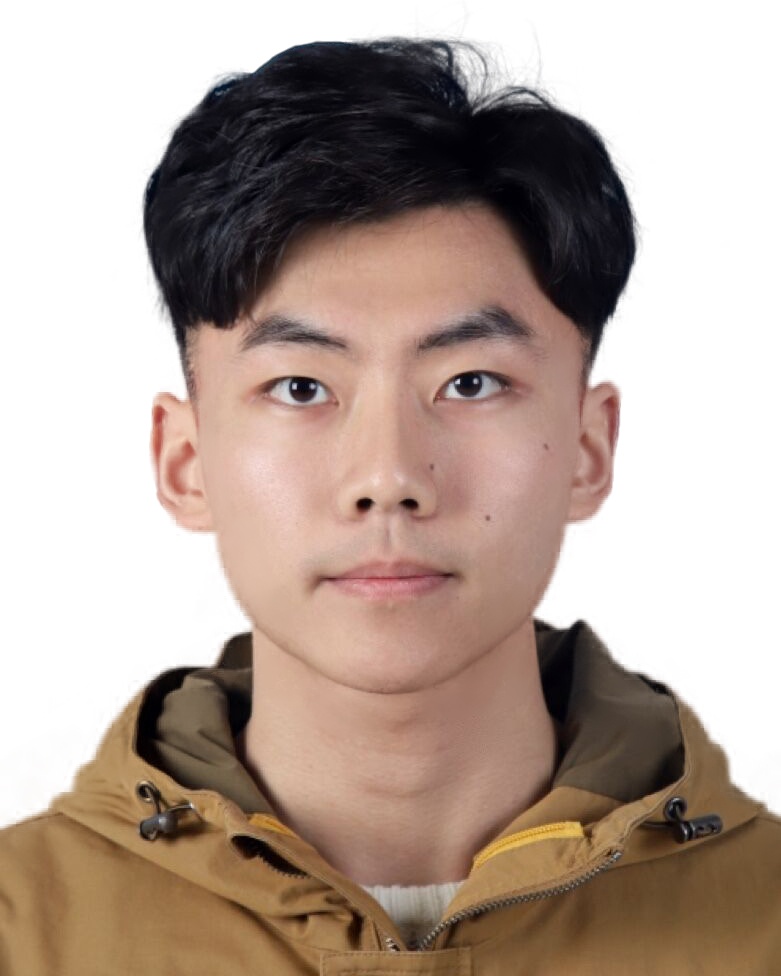}}]{Fangcheng Zhu}
    received the B.E. degree in automation in 2021 from the School of Mechanical Engineering and Automation, Harbin Institute of Technology, Shenzhen, China. He is currently working toward the Ph.D. degree in robotics with the Department of Mechanical Engineering, University of Hong Kong (HKU), Hong Kong, China.
    His research interests include LiDAR-based simultaneous localization and mapping (SLAM),sensor calibration, and aerial swarm systems.
\end{IEEEbiography}

\vspace{-60pt}
\begin{IEEEbiography}[{\includegraphics[width=1in,height=1.25in,clip,]{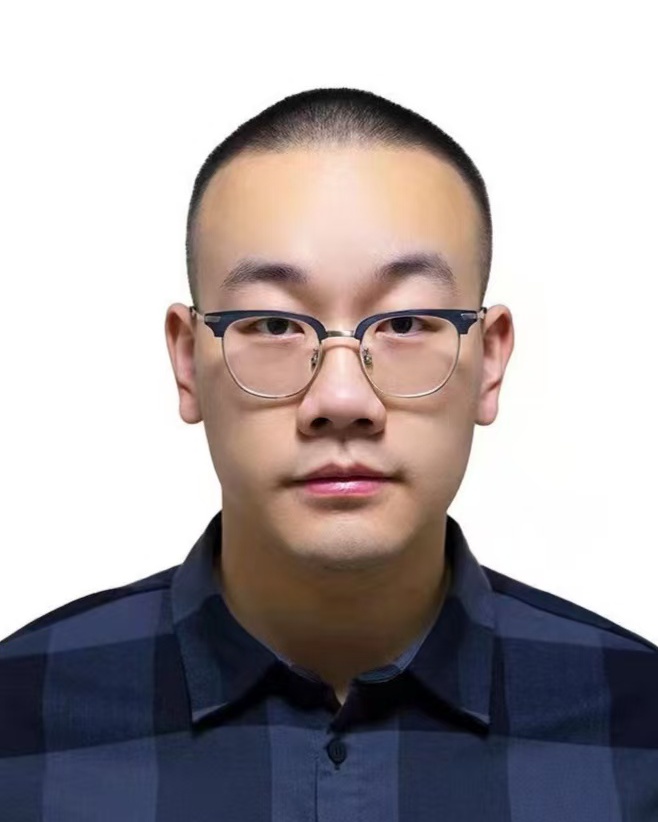}}]{Longji Yin}
    Longji Yin received the B.Eng. degree in automation and the B.A. degree in english literature from Zhejiang University, Zhejiang, China, in 2019 and the M.Sc. degree in robotics from the Johns Hopkins University, Maryland, U.S. in 2021.  He is currently working toward the Ph.D. degree in mechanical engineering at the University of Hong Kong, Hong Kong.
    His research interests include motion planning and control for autonomous aerial robots.
\end{IEEEbiography}

\vspace{-60pt}
\begin{IEEEbiography}[{\includegraphics[width=1in,height=1.25in,clip,]{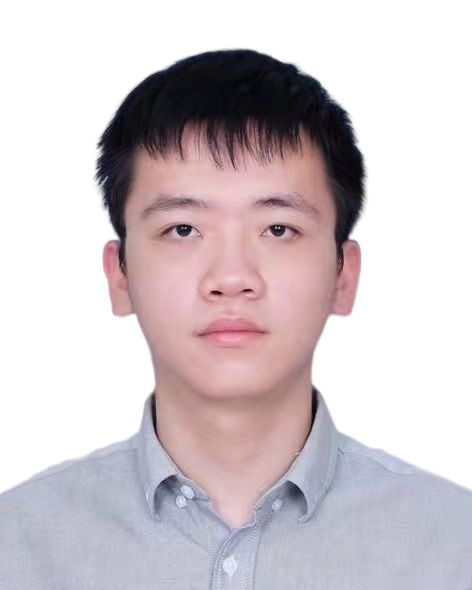}}]{Fanze Kong}
    received the Ph.D. degree in Mechanical Engineering with the Department of the University of Hong Kong, Hong Kong, China and received the B.S. degree in Flight Vehicle design and engineering from Harbin Institute of Technology, Harbin, China, in 2020. His research interests include UAV design, swarm UAV autonomous navigation and LiDAR simulation.
\end{IEEEbiography}

\vspace{-60pt}
\begin{IEEEbiography}[{\includegraphics[width=1in,height=1.25in,clip,]{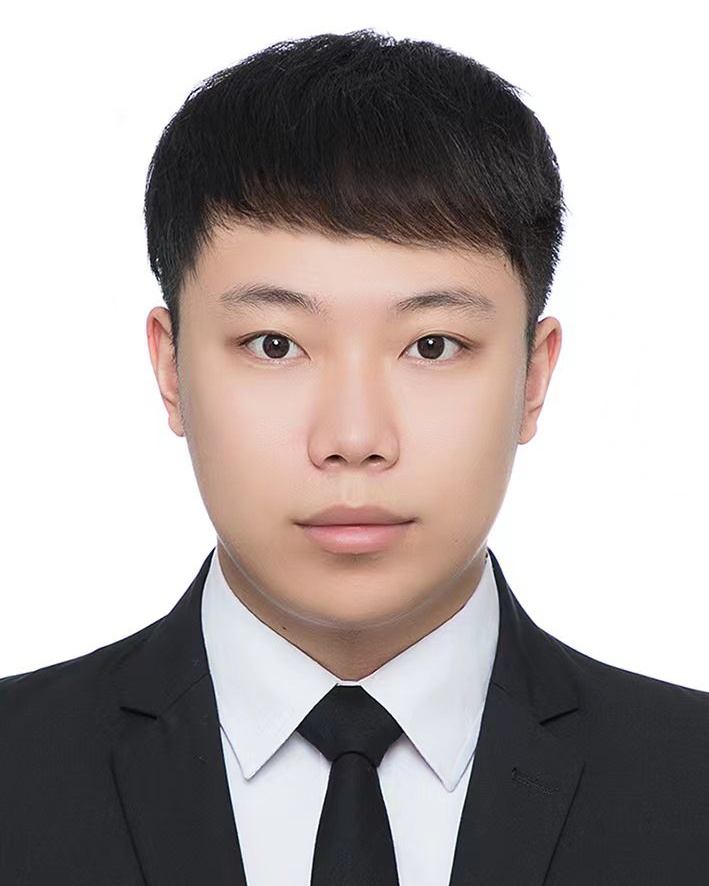}}]{Rundong Li} is a PhD candidate in the Department of Mechanical Engineering at The University of Hong Kong. He received his Bachelor’s degree in Robotics from Beihang University in 2022. His research interests include large-scale multi-sensor bundle adjustment and mapping.
\end{IEEEbiography}

\vspace{-60pt}
\begin{IEEEbiography}[{\includegraphics[width=1in,height=1.25in,clip,]{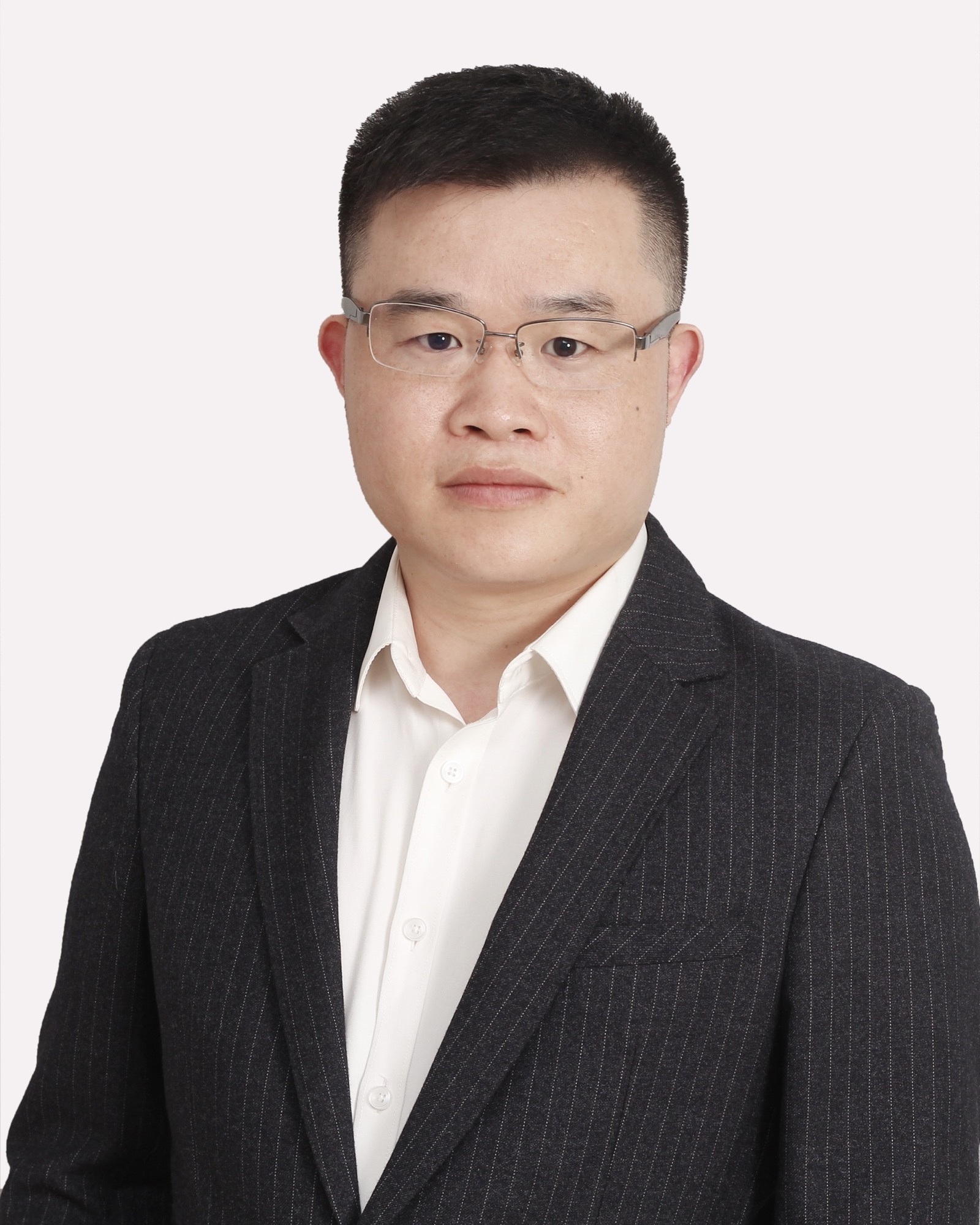}}]{Fu Zhang} received his B.E. degree in Automation from the University of Science and Technology of China (USTC), Hefei, Anhui, China, in 2011, and the Ph.D. degree in Controls from the University of California, Berkeley, CA, USA, in 2015. His Ph.D. work focused on self-calibration and control of micro rate-integrating gyro sensors. In 2016, Dr. Zhang shifted his research to the design and control of Unmanned Aerial Vehicles (UAVs) as a Research Assistant Professor in the Robotics Institute of the Hong Kong University of Science and Technology (HKUST). He joined the Department of Mechanical Engineering, the University of Hong Kong (HKU), as an Assistant Professor from Aug 2018. His current research interests are on robotics and controls, with focus on UAV design, navigation, control, and lidar-based simultaneous localization and mapping.
\end{IEEEbiography}

\end{document}